%% file: paper.tex
\documentclass[journal]{vgtc}                     

\onlineid{1130}

\vgtccategory{Research}

\vgtcpapertype{Application}

\title{CPTracker : Hypervelocity Impact Tracker with Topological Data Analysis  }
\title{DebrisTracer:\\ Robust Temporal Tracking
in Hypervelocity Impact Acquisitions}
\title{DebrisTracer: Reliable Tracking in Hypervelocity Impact Fast Imaging}
\author{%
  Théophane Loloum,
  Fabien Vivodtzev, 
  David Hébert,
  Baptiste Reynier, 
  Michel Arrigoni,
  and
  Julien Tierny
}

\authorfooter{
  \item Théophane Loloum, Fabien Vivodtzev, Baptiste Reynier, and  David Hébert are with the CEA.
  E-mail: \{firstname.lastname\}@cea.fr 
  \item Michel Arrigoni is with ENSTA.
  E-mail: \{firstname.lastname\}@ensta.fr
  \item Julien Tierny is with the CNRS and Sorbonne University.
  E-mail: \{firstname.lastname\}@sorbonne-universite.fr
}

\input{abstract.tex}

\graphicspath{{figs/}{figures/}{pictures/}{images/}{./}}

\usepackage{tabu}                      
\usepackage{booktabs}                  
\usepackage{lipsum}                    
\usepackage{mwe}                       
\usepackage{mathptmx}                  
\usepackage{amsmath}
\usepackage{amssymb}
\usepackage{url}                       
\usepackage{xr}
\usepackage{calrsfs}
\DeclareMathAlphabet{\pazocal}{OMS}{zplm}{m}{n}

\usepackage{mathptmx}                  

\makeatletter
\@ifundefined{KV@Gin@alt}{\define@key{Gin}{alt}[]{}}{}
\makeatother

\begin{document}



\input{notations.tex}

\input{introduction.tex}

\input{context.tex}

\input{background.tex}

\input{tracking.tex}

\input{physics.tex}

\input{results.tex}

\input{conclusion.tex}

\clearpage

\clearpage

\acknowledgments{%
\scriptsize{%
This work is partially supported by the European Commission
grant ERC-2019-COG \emph{``TORI''} (ref. 863464,
\url{https://erc-tori.github.io/}).}
}

\bibliographystyle{abbrv-doi-hyperref}

\bibliography{paper}

\appendix 

\input{parameterSetting.tex}

\input{laser.tex}

\input{gallery.tex}

\end{document}

%% file: abstract.tex
\abstract{Abstract, line 1

  line 2

  line 3

  line 4

  line 5

  line 6

  line 7

  line 8

  line 9

  line 10
}

\abstract{This application paper presents \emph{DebrisTracer}, a framework for the reliable tracking of debris in hypervelocity impact fast imaging. These
noisy and highly specific
datasets capture the ejection of a large number of debris
fragments
after the impact of a projectile launched at hypervelocity  into
a target material.
The reliable estimation of 
debris mass and speed distributions
is of major importance in aerospace applications.
We document how to extend an off-the-shelf topology tracking framework \cite{soler_ldav18} based on critical point extraction and matching, in order to incorporate domain knowledge and physical assumptions. Our approach automatically produces an accurate and reliable debris tracking,
enabling
an interpretable
visual
analysis of this complex space-time phenomenon. Extensive experiments demonstrate the accuracy improvements provided by our approach over
established tools used by domain experts 
\cite{tinevez2017trackmate, Ershov2022} in terms of physical validation, specifically via
the prediction
of the experimental ejected mass and crater depth profiles.
We illustrate the utility of our approach across several use cases (with varying impact angles and physics). We show that our statistical summaries enable the
visual
identification of distinct regimes within the debris population,
corroborating
and refining
prior expectations of domain experts.
Our
database and our C++ implementation are available at this address: \href{https://github.com/tloloum/DebrisTracer}{https://github.com/tloloum/DebrisTracer}.}

\keywords{Hypervelocity impact, feature tracking, topological data analysis.}

\teaser{
  \centering
  \includegraphics[width=\linewidth]{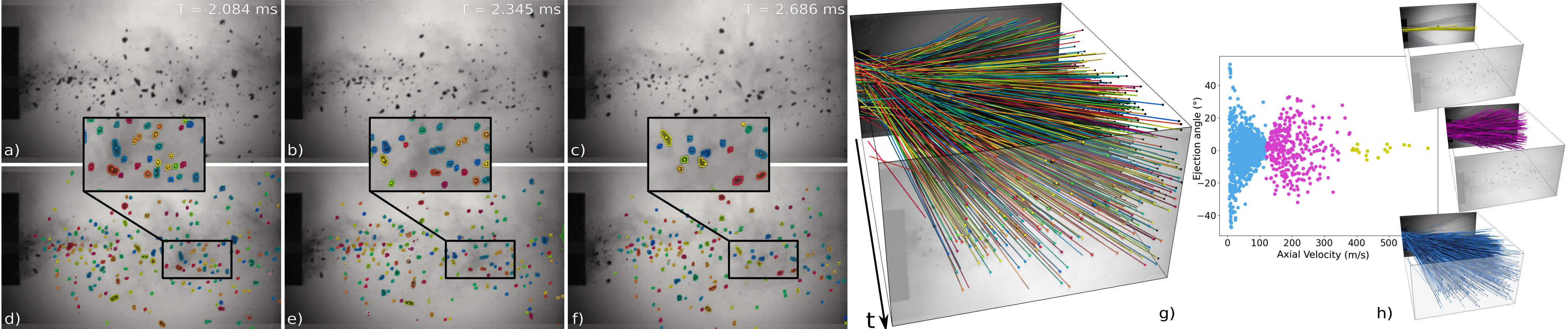}
  \caption{In experimental hypervelocity impacts, a projectile is launched at
  \todo{5 km/s} to
  the material under study
  (left, black rectangle), e.g., to emulate satellite collisions. This produces
  many
  debris fragments, of distinct masses and speeds \emph{(a)-(c)}. The reliable estimation of
  these
  mass and speed distributions is of major importance in aerospace design.
  \emph{DebrisTracer} extends an off-the-shelf topology tracking approach \cite{soler_ldav18} to incorporate domain knowledge and physical assumptions.
  It
  automatically produces an accurate and reliable debris tracking \emph{(d)-(g)}, supporting  interpretable
  visual
  analyses,
  facilitating
  the understanding of this complex
  space-time
  phenomenon. In this experiment, it enables
  distinguishing and exploring
  three debris
types
  \emph{(h)}: fast (yellow), medium (purple) and slow (blue)
  ejections.
  }
  \label{fig:teaser}
}

%% file: notations.tex
\renewcommand{\mathcal}[1]{\pazocal{#1}}

\newcommand{\electronDensity}{\rho}

\newcommand{\surface}{S}

\newcommand{\frameNumber}{n_t}
\newcommand{\trajectoryNumber}{n_T}

\newcommand{\trajectory}{P}
\newcommand{\trajectories}{\mathcal{P}}
\newcommand{\criticalPoint}{m}
\newcommand{\timeMap}{t}
\newcommand{\velocityVector}{v}
\newcommand{\scalingVector}{s}
\newcommand{\velocityLine}{l}
\newcommand{\trackingSegment}{S}
\newcommand{\segmentSequence}{\mathcal{S}}
\newcommand{\myangle}{\theta}
\newcommand{\trajectoryLine}{L}
\newcommand{\lineCollection}{\mathcal{L}}
\newcommand{\debrisShape}{D}
\newcommand{\regionArea}{a}
\newcommand{\debrisArea}{A}
\newcommand{\debrisRadius}{r}
\newcommand{\debrisVolume}{V}
\newcommand{\debrisMass}{M}
\newcommand{\destructionCost}{\omega}
\newcommand{\coneAngle}{\Theta}
\newcommand{\persistenceThreshold}{\epsilon_{min}}
\newcommand{\temporalGap}{\delta}

\newcommand{\extremumGraph}{\mathcal{E}}
\newcommand{\minimumSet}{\mathcal{N}}
\newcommand{\unstableSets}{\mathcal{A}}
\newcommand{\occurrence}{\mathcal{O}}

\newcommand{\signal}{\mathcal{S}}
\newcommand{\noise}{\mathcal{N}}
\newcommand{\simplex}{\sigma}
\newcommand{\domain}{\mathcal{K}}
\newcommand{\numberOfVertices}{n_v}
\newcommand{\numberOfSimplices}{n_\simplex}
\newcommand{\dataVector}{v}
\newcommand{\dataVectorSpace}{\mathcal{V}}
\newcommand{\filtration}{\mathcal{F}}
\newcommand{\persistenceMap}{\mathcal{P}}
\newcommand{\energy}{\mathcal{E}}
\newcommand{\loss}{\mathcal{L}}
\newcommand{\range}{\mathbb{R}}
\newcommand{\sublevelset}[1]{#1^{-1}_{-\infty}}
\newcommand{\superlevelset}[1]{#1^{-1}_{+\infty}}
\newcommand{\Star}{St}
\newcommand{\Link}{Lk}
\newcommand{\diagram}{\mathcal{D}}
\newcommand{\target}{\diagram_T}
\newcommand{\complexity}{\mathcal{O}}

\newcommand{\face}{\tau}
\newcommand{\lowerlink}{\Link^{-}}
\newcommand{\upperlink}{\Link^{+}}
\newcommand{\Index}{\mathcal{I}}
\newcommand{\offset}{o}
\newcommand{\Natural}{\mathbb{N}}
\newcommand{\criticalSet}{\mathcal{C}}

\newcommand{\wasserstein}[1]{\mathcal{W}_{#1}}
\newcommand{\projection}{\Delta}
\newcommand{\hierarchy}{\mathcal{H}}
\newcommand{\decimation}{D}
\newcommand{\xDimD}{L_x^\decimation}
\newcommand{\yDimD}{L_y^\decimation}
\newcommand{\zDimD}{L_z^\decimation}
\newcommand{\xDim}{L_x}
\newcommand{\yDim}{L_y}
\newcommand{\zDim}{L_z}
\newcommand{\cubicalComplex}{\mathcal{C}}
\newcommand{\Grid}{\mathcal{G}}
\newcommand{\GridD}{\mathcal{G}^\decimation}
\newcommand{\x}{\phantom{x}}
\newcommand{\Mod}{\;\mathrm{mod}\;}
\newcommand{\NN}{\mathbb{N}}
\newcommand{\forwardIntegralLine}{\mathcal{L}^+}
\newcommand{\backwardIntegralLine}{\mathcal{L}^-}
\newcommand{\triangulationOp}{\phi}
\newcommand{\decimationOp}{\Pi}
\newcommand{\isovalue}{w}
\newcommand{\persistence}{p}
\newcommand{\pointMetric}{d}
\newcommand{\diagramSet}{\mathcal{S}_\mathcal{D}}
\newcommand{\diagramSpace}{\mathbb{D}}
\newcommand{\jointree}{\mathcal{T}^-}
\newcommand{\splittree}{\mathcal{T}^+}
\newcommand{\mergetree}{\mathcal{M}}
\newcommand{\tree}{\mergetree}
\newcommand{\depth}{d}
\newcommand{\mergetreeSet}{\mathcal{S}_\mathcal{T}}
\newcommand{\branchset}{\mathcal{S}_\mathcal{B}}
\newcommand{\branchspace}{\mathbb{B}}
\newcommand{\mergetreeSpace}{\mathbb{T}}
\newcommand{\editdistance}{D_E}
\newcommand{\wassersteinTree}{W^{\mergetree}_2}
\newcommand{\distanceSequence}{d_S}
\newcommand{\branchtree}{\mathcal{B}}
\newcommand{\branchtreeSet}{\mathcal{S}_\mathcal{B}}
\newcommand{\branchtreeSpace}{\mathbb{B}}
\newcommand{\forest}{\mathcal{F}}
\newcommand{\sequenceSpace}{\mathbb{S}}
\newcommand{\forestMatrix}{\mathbb{F}}
\newcommand{\treeMatrix}{\mathbb{T}}
\newcommand{\normalizedLocation}{\mathcal{N}}
\newcommand{\normalizedWasserstein}{W^{\normalizedLocation}_2}
\newcommand{\geodesictree}{\mathcal{G}}
\newcommand{\dummyVector}{\mathcal{V}}
\newcommand{\geodesictreeVec}{g}
\newcommand{\geodesicAxis}{\mathcal{A}}
\newcommand{\directionVector}{\mathcal{V}}
\newcommand{\geodesicdiagram}{\mathcal{G}^{\diagram}}
\newcommand{\reconstructionError}{E_{L_2}}
\newcommand{\pcaBasis}{B_{\mathbb{R}^d}}
\renewcommand{\pcaBasis}{B}
\newcommand{\origin}{o_b}
\newcommand{\sizeEncoding}{n_e}
\newcommand{\sizeDecoding}{n_d}
\newcommand{\linearTransformation}{\psi}
\newcommand{\unitTransformation}{\Psi}
\renewcommand{\origin}{o}
\newcommand{\bdtOrigin}{\mathcal{O}}
\newcommand{\activation}{\sigma}
\newcommand{\validBDT}{\gamma}
\newcommand{\mtPgaBasis}{B_{\branchtreeSpace}}
\newcommand{\mtPgaError}{E_{\wassersteinTree}}
\newcommand{\frechetEnergy}{E_F}
\newcommand{\geodesicExtremity}{\mathcal{E}}
\newcommand{\vectorNotation}[1]{\protect\vv{#1}}
\renewcommand{\vectorNotation}[1]{#1}
\newcommand{\axisNotation}[1]{\protect\overleftrightarrow{#1}}
\newcommand{\individualEnergy}{E}
\newcommand{\ensembleSize}{N}
\newcommand{\numberBranchinBarycenter}{N_1}
\newcommand{\numberGeodesicSamples}{N_2}
\newcommand{\planarGridX}{N_x}
\newcommand{\planarGridY}{N_y}
\newcommand{\regularGrid}{G}
\newcommand{\distanceMatrix}{\mathbb{D}}
\newcommand{\maxDimensions}{{d_{max}}}
\newcommand{\projectionOperator}{\mathcal{P}}
\newcommand{\reconstructed}[1]{\widehat{#1}}
\newcommand{\gt}{>}
\newcommand{\lt}{<}
\newcommand{\branch}{b}
\newcommand{\nonLinearFunction}{\sigma}
\newcommand{\batchSequence}{S}
\newcommand{\homologyGroup}{\mathcal{H}}
\newcommand{\bettiNumber}{\beta}
\newcommand{\still}{\mathcal{S}}

\newcommand{\weight}{w}

\newcommand{\julien}[1]{\textcolor{blue}{#1}}
\renewcommand{\julien}[1]{\textcolor{black}{#1}}

\newcommand{\fv}[1]{\textcolor{blue}{#1}}
\renewcommand{\fv}[1]{\textcolor{black}{#1}}

\newcommand{\theophane}[1]{\textcolor{magenta}{#1}}
\renewcommand{\theophane}[1]{\textcolor{black}{#1}}

\newcommand{\david}[1]{\textcolor{olive}{#1}}
\renewcommand{\david}[1]{\textcolor{black}{#1}}

\newcommand{\gosia}[1]{\textcolor{purple}{#1}}
\renewcommand{\gosia}[1]{\textcolor{black}{#1}}

\newcommand{\revision}[1]{\textcolor{black}{#1}}
\newcommand{\minor}[1]{\textcolor{blue}{#1}}

\newcommand{\discuss}[1]{\textcolor{black}{#1}}

\renewcommand{\figureautorefname}{Fig.}
\renewcommand{\sectionautorefname}{Sec.}
\renewcommand{\subsectionautorefname}{Sec.}
\renewcommand{\equationautorefname}{Eq.}
\renewcommand{\tableautorefname}{Tab.}
\newcommand{\algorithmautorefname}{Alg.}
\newcommand{\lineautorefname}{Alg.}

\newcommand{\todo}[1]{\textcolor{red}{#1}}
\renewcommand{\todo}[1]{\textcolor{black}{#1}}

\newcommand{\todoRev}[1]{\textcolor{red}{#1}}

\newcommand{\mycaption}[1]{
\caption{#1}
}

\newcommand{\myparagraph}[1]{
\noindent\textbf{#1}}

\newcommand{\journal}[1]{\textcolor{blue}{#1}}
\renewcommand{\journal}[1]{\textcolor{black}{#1}}

\newcommand{\myspace}{\vspace{-0.0775ex}}

\newcommand{\myspaceFour}{\vspace{-0.35ex}}



%% file: introduction.tex
\firstsection{Introduction}

\maketitle
\begin{figure}
  \centering
  \includegraphics[width=\linewidth]{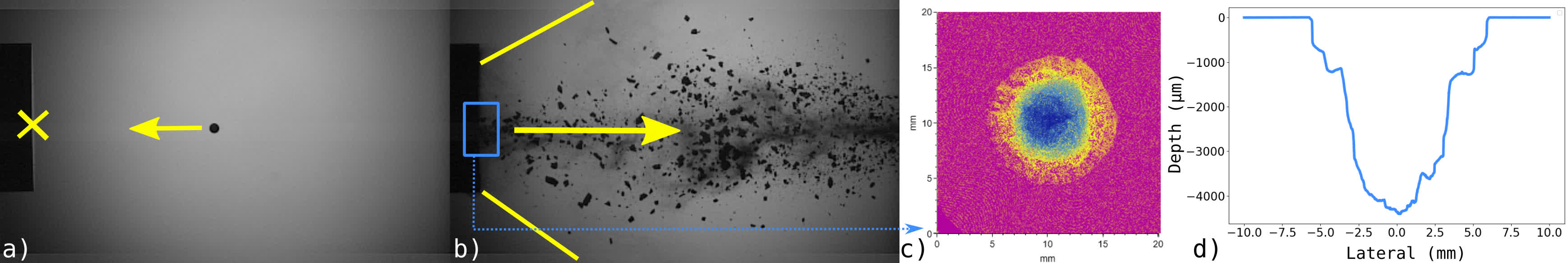}
  \caption{Experimental setup for HVI acquisitions: a projectile \emph{(a)} is launched to the target (cross) at a typical speed
  of
  \todo{5~km/s}.
  This results in a cone-shaped ejection of
many
  debris fragments \emph{(b)}. The impact leaves a
  crater on the target, captured by a laser scan \emph{(c)}, enabling the modeling of crater
  depth
  profiles \emph{(d)} along the target segment.
  }
  \label{fig:applicationContext}
\end{figure}

The study of \david{HyperVelocity Impacts (HVI)} is of major importance in
material science,
with applications in aerospace design (e.g., to emulate satellite collisions \cite{MMOD1, MMOD2}), planetary defense (to study asteroid deflection \cite{Cheng2018, Jutzi2014}),
or high energy density physics
(for shielding in
confinement fusion \cite{NIFLMJ, NIF}).
Upon such impacts,
the material under study undergoes a process called \emph{shock-wave compression}, leading to
the rapid formation of a crater, as well as the
high-speed
ejection of debris.
In particular, the mass and velocity distributions of these debris are critical for modeling secondary damage, debris cloud evolution, shielding degradation or
trajectory
perturbations.
Therefore, it is crucial for 
aerospace
engineers to
obtain accurate estimates
of these distributions. For that, \david{several diagnostics can be used, among which} high-frame-rate videos \david{of HVI experiments performed in laboratory conditions \cite{HEBERT2022}}  (\autoref{fig:applicationContext}).
However, the
resulting
datasets are particularly challenging to analyze, and
existing
frameworks
used by domain experts
\cite{tinevez2017trackmate, Ershov2022}
often
provide sub-optimal results, due to the specifics of
HVI.

Specifically, given the technical complexity and costs of
HVI
controlled experiments,
only few datasets are available, preventing the reliable training of machine learning models. Moreover, for these very specific datasets, general-purpose segmentation models \cite{sam2} (trained on generic images)
exhibit identification rates that are both too low and inconsistent across consecutive frames
(\autoref{fig:dataChallenge}).
These failures can be partly explained by
the challenges
of
HVI
imaging (\autoref{sec_domainInfo}). For instance,
due to the high-frame-rate camera setups,
these acquisitions
are typically affected by
strong
vignetting artifacts,
resulting in
large contrast and brightness variations across the image. Specifically, these variations
make it difficult to segment debris
based on pixel intensity thresholding (\autoref{fig:dataChallenge}).
This  challenges \emph{(i) debris characterization}.
In addition,
for fast debris,
the acquisition frequency of the employed cameras may be insufficient, leading to large
displacements for a single debris fragment between consecutive frames.
Moreover,
due to occlusion,
several
debris may overlap in the image
while they are distinct objects in 3D.
These
two difficulties
confound
tracking approaches based on object overlap detection, challenging \emph{(ii) debris tracking}.

Overall, the above two issues challenge the existing tracking approaches  which use intensity thresholding for object characterization and overlap estimation for object tracking.
In topological data analysis \cite{edelsbrunner09}, several methods
leveraged
\emph{(i)} persistent homology and
\emph{(ii)}
transportation metrics
to address these respective issues,
as documented in recent surveys and benchmarks \cite{surveyComparison2021, emmaBenchmark22, LeThanh25}. While a subset of these methods focus on tracking of complex structural patterns (e.g., based on merge trees), the binary characterization of objects in our application (debris or background) allows for a more straightforward and less constrained strategy.
Specifically, we focus in this paper on a simple pre-existing approach \cite{soler_ldav18} based on critical point extraction and tracking, and
discuss
its limitations for our application.
We also document the process of extending this off-the-shelf topology tracking framework to incorporate the necessary domain knowledge  (\autoref{sec_topoTracking})  and physical
assumptions
(\autoref{sec_physics}) to improve its accuracy and usability,
while supporting trustworthiness and interpretability for domain experts.
Extensive experiments (\autoref{sec_results}) demonstrate the accuracy improvements, both in terms of debris characterization and physical validation (via the estimated ejected mass and reconstructed crater profiles), over  established tools  used by domain experts \cite{tinevez2017trackmate, Ershov2022} and over our selected off-the-shelf topology tracking framework \cite{soler_ldav18}. Our improved tracking offers new capabilities for analyzing the main trends in debris types, yielding new
visual
insights for experts, as documented in several use cases (\autoref{sec_caseStudies}), studying several impact angles and physics (projectile versus laser).
We provide in additional material our C++ implementation, along with our database of hypervelocity impact acquisitions, which may constitute a benchmark for future work in feature tracking.

\subsection{Related work}
\label{sec_relatedWork}

This section reviews the literature related to our
work,
which can be classified into the following two categories.

\noindent
\textbf{\emph{(i)} High velocity impact analysis:}
With the development of the aerospace industry since the twentieth century, hypervelocity impacts (HVI) have received significant attention in material science and reference textbooks are available
\cite{kinslow1970high}.
\david{Experimental
\julien{protocols}
have been proposed to
assess
the size and/or velocity distribution of ejecta\julien{, based on devices in charge of
physically
collecting the debris (e.g., with gel or aerogels soft-recovery collectors \cite{lescoute2012}).}
In some cases,
gravity can be used to separate the ejecta and collect them into bins \cite{MICHIKAMI2007}. Other authors analyze the secondary craters observed on witness plates \cite{Nishida}. However, these methods sometimes require dedicated experiments to ensure a good knowledge of the collector material behavior under impact \cite{POLYCARBONATE, paraffin}.
Because of the difficulty to extract both size and velocity distributions from a single diagnostic, recent experiments have shown the value of combining different diagnostics \cite{Reynier,Nakamura,Yamaguchi}.}
Another line of experiments aim at providing ``in-flight'' diagnostics.
\david{For instance, Photonic Doppler Velocimetry (PDV) allows measurement of multiple fragments velocities with a laser probe \cite{Prudhomme_2014}.
Acquisitions based on high-speed cameras are becoming widely used in HVI experiments. However, the
quantitative extraction of debris distributions from such videos remains
challenging}, as it requires a reliable frame-to-frame tracking method, for capturing the trajectories of the individual debris fragments.
A few authors
proposed custom tracking  approaches \cite{WATSON2017,Matura}, while
several
studies
rely on
established tracking frameworks  \cite{ghosh2025quantifying, eidevag2021collisional}.
In particular,
\revision{the \emph{Fiji/ImageJ} \cite{schindelin2012fiji} plugin}
\emph{TrackMate} \cite{tinevez2017trackmate, Ershov2022}, 
\revision{provides a generic}
framework
for particle tracking in 2D videos. \revision{It} is considered as a
\revision{go-to}
key tool by
HVI
experts\revision{, used 
frequently in the literature, e.g., \cite{ghosh2025quantifying, 
eidevag2021collisional}}.
It implements a variety of algorithms for particle characterization (intensity thresholding, derivative analysis, etc.) and tracking (by overlaps or
optimal assignment, etc.).
However, to our knowledge, \emph{TrackMate} does not support topological methods for characterization and tracking. \autoref{sec_physicalValidation}
compares
\emph{TrackMate} and our framework,
demonstrating substantial gains for our approach, in terms of
physical validation.


\noindent
\textbf{\emph{(ii)} Topology tracking:}
Topological methods \cite{edelsbrunner09} have considerably developed over the last two decades in data
visualization \cite{heine16}, for the robust and multi-scale representations of structural patterns in complex data. They rely on a family of concise topological descriptors, such as the persistence diagram
\cite{edelsbrunner02}, the merge and contour trees \cite{carr00}, Reeb graphs \cite{parsa12} or Morse-Smale complexes \cite{ShivashankarN12}. These descriptors have been successfully applied in a variety of applications,
including
combustion \cite{bremer09},
material science \cite{soler_ldav19},
fluid dynamics \cite{nauleau_ldav22},
bioimaging \cite{topoAngler}, astrophysics \cite{shivashankar2016felix} or
chemistry \cite{daniel_vis25}. Since topological methods have a strong focus on structure rather than geometry, they naturally appear as promising candidates for characterizing and tracking debris, which are disconnected pieces exhibiting a wide variety of shapes.

\begin{figure}
  \centering
  \includegraphics[width=\linewidth]{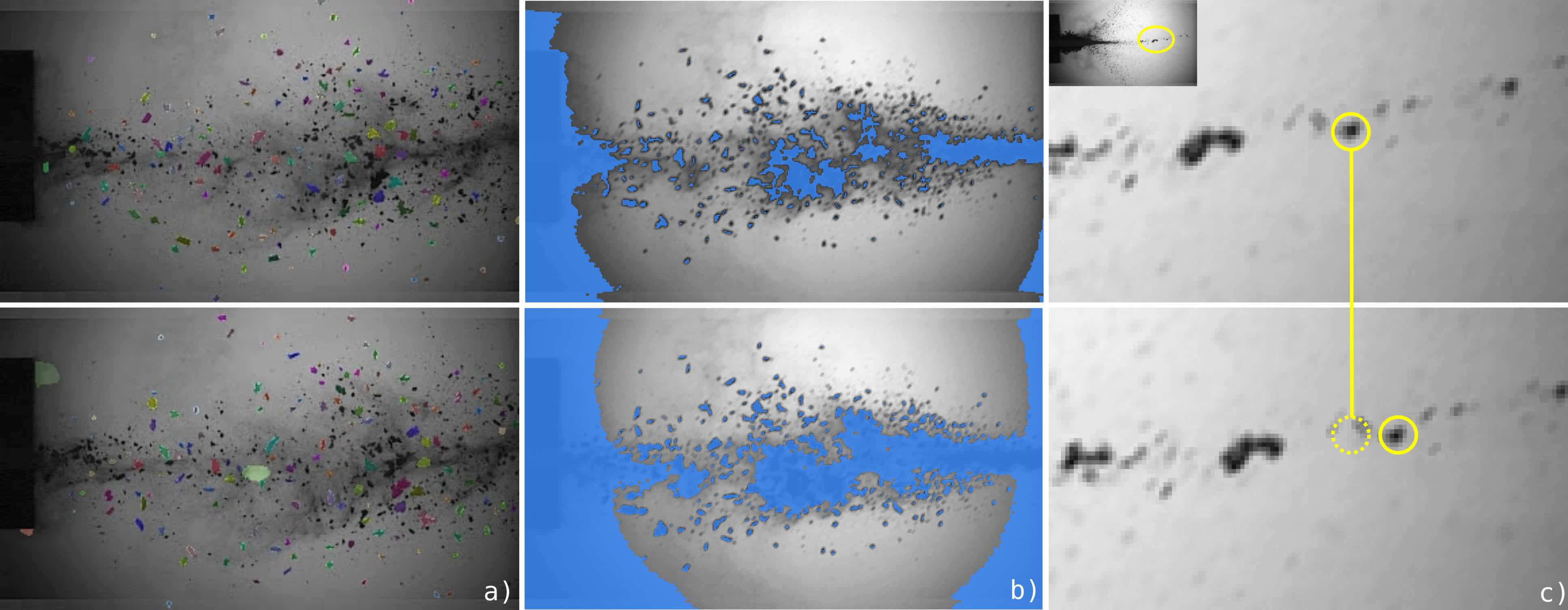}
  \caption{Challenges of HVI acquisitions: given the specifics of this imaging modality (noise, vignetting artifacts, occlusion, debris shape variability), general-purpose segmentation models \cite{sam2} provide correct classification rates that are low \emph{(a)} and
  inconsistent
  across time steps (top versus bottom). Vignetting
  artifacts
challenge automatic
thresholding
(blue)
for debris characterization \emph{(b)}, as a given threshold may
both
under-segment (i.e., capture multiple debris within a single connected component), and
  miss certain debris, irrespective of the selected isovalue.
   Insufficient
   frame-rates
   result in debris which do not overlap in consecutive time steps \emph{(c)}, challenging overlap-based tracking approaches.}
  \label{fig:dataChallenge}
\end{figure}

Several approaches have been
proposed
for tracking particles in simulated vector field data \cite{featureFlowFields, stableFeatureFlowFields, combinatorialFeatureFlowFields, ftk}. However, since the input data is provided in our work in the form of gray-scale 2D videos, we will focus in the remainder on tracking approaches for scalar data.

In scalar data analysis, several methods have been designed based on an object segmentation (e.g.,  using merge or contour trees), and relying on overlap detection between consecutive time steps to perform tracking
\cite{bremer09, Wathsala12,
jonas20, oesterling17, SaikiaW17, Saikia20,
andrea20}. However,
HVI
images challenge debris characterization based on intensity thresholding due to important vignetting artifacts (\autoref{fig:dataChallenge}, \autoref{sec_domainInfo}). Moreover,
insufficient
frame-rates
may prevent debris overlap over consecutive time steps, challenging overlap-based approaches.

Alternative strategies for object characterization rely on
topological persistence
\cite{edelsbrunner02}, which can be interpreted as a robust importance measure, enabling the identification of salient features, irrespective of vignetting artifacts (see \autoref{sec_characterization}).
In this framework, feature tracking is no longer estimated via overlap detection, but by
optimizing an assignment between the topological descriptors of consecutive time steps.
This overall strategy has been employed in a variety of application domains, including
fluid dynamics \cite{soler_ldav19}, cloud tracking \cite{harishClouds, liTopoInVis25} or cyclone tracing \cite{emmaCyclone, vijayCyclone, ingridCycloneTopoInVis19}.
We refer the reader to recent surveys
for further details on this overall strategy \cite{surveyComparison2021, LeThanh25,emmaBenchmark22}.

Nilsson et al. introduce an approach for matching extrema in time-varying scalar fields \cite{emmaGradientBased},
based on gradient path integration. However, the local nature of gradient and the cumulative error of its integration may be challenged
by fast moving objects, in particular in the presence of noise.
Soler et al. introduce an assignment optimization approach between persistent
extrema
\cite{soler_ldav18}, combining criteria based on persistence and distances in the
original domain.
Several assignment
approaches have been considered for similarity estimation and tracking between merge trees \cite{BeketayevYMWH14, tinoEuroVis14, SridharamurthyM20, pont_vis21,
florian23},
extremum graphs \cite{das24}, or Reeb graphs \cite{timeReeb, Weber2011,
reebEditDistance}. These
methods
enable the tracking of complex structural patterns, linking the features of interest together. However, our application setup (\autoref{sec_applicationContext}) only requires a binary characterization of objects: pixels are either debris or background.
The  debris cloud does not exhibit
specific local structures
between the
individual debris fragments,
that would need to be characterized and tracked with advanced topological descriptors,
such as
merge trees.
Debris fragments simply need to be isolated and tracked.
For this reason, we focus in our work on a simpler approach based on
extremum
extraction and tracking \cite{soler_ldav18}. Specifically, it matches
extrema
based on an assignment optimization, irrespective of the structural patterns
they may form locally or globally.
This allows for a straightforward and less constrained matching strategy. Then,
this
paper
describes
how to extend this topology tracking framework to incorporate the necessary domain knowledge about the acquisitions (\autoref{sec_topoTracking}) and physical assumptions (\autoref{sec_physics}) to improve its accuracy, trustworthiness and interpretability for domain experts.

\subsection{Contributions}
This paper makes the following new contributions:
\begin{enumerate}
 \item
 \emph{Domain-tailored approach for debris tracking:} We document how to extend an off-the-shelf topology tracking framework \cite{soler_ldav18} to incorporate domain knowledge
 and physical
 assumptions
 to improve tracking accuracy and usability. We document the resulting improvements both in terms of debris characterization and physical validation over established domain tools \cite{tinevez2017trackmate, Ershov2022}.
 \revision{While our approach builds on an existing framework \cite{soler_ldav18},
 we followed a general methodology
 to adapt it
 to our application constraints, based on:
 \emph{(i)}
 a formalization of observations, assumptions and physical hypotheses (\autoref{sec_domainInfo}),
 \emph{(ii)} a specific modification of the existing framework \cite{soler_ldav18}  to incorporate domain knowledge (\autoref{sec_topoTracking}) and
 \emph{(iii)} a novel, tailored post-processing of the tracking output to integrate physical assumptions (\autoref{sec_physics}). We believe that this generic, three-step workflow may be applied to other domain-specific applications, particularly those involving feature tracking. In that regard, our work provides an example of how to implement this methodology, which may be useful for other researchers
 applying
 topological methods to domain sciences.}
 \item \emph{Case studies:} The tracking performance of our approach allows for novel capabilities for the reliable visual analysis of debris, in terms of mass and velocity distributions. We illustrate these features on several use cases, for varying impact angles and physics.
 Our analysis corroborates and refines prior hypotheses from domain experts regarding the main trends in debris distributions. \revision{This hypothesis refinement was made possible by the visual representations (angle/velocity scatterplots, mass/velocity distributions) generated by our approach, illustrating a concrete case of applied end-to-end data transformation for
 visual reasoning.}
 \item \emph{Implementation:} We provide a C++ implementation of our approach that can be used for reproducibility purposes.
 \item \emph{Database:} We contribute our database of hypervelocity impact acquisitions,
 i.e., \todo{8} two-dimensional time-varying scalar fields (about \todo{2.5} gigabytes),
 counting between \todo{256 and 1050} frames, capturing different dynamic fragmentation modalities (projectile-based and laser-based).
 This database may constitute a benchmark for future research in feature tracking.
\end{enumerate}

%% file: context.tex
\section{Application context}
\label{sec_applicationContext}

\begin{figure*}
  \centering
  \includegraphics[width=\linewidth]{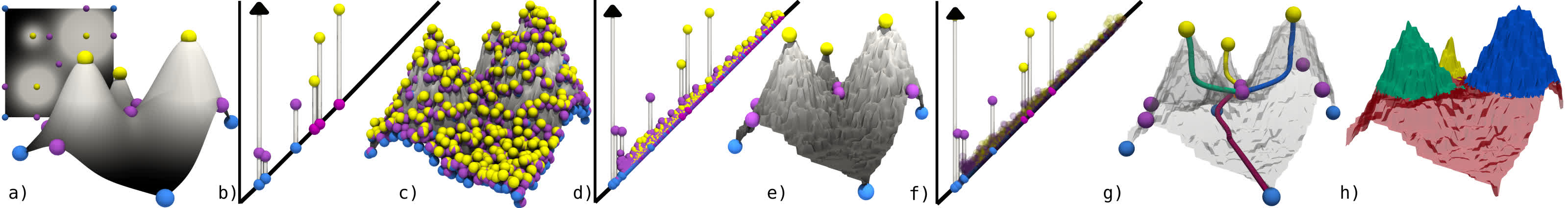}
	\caption{Multiscale topological analysis pipeline used in our work: the topological features of a scalar field \emph{(a)} can be captured by pairs of critical points (i.e., \emph{persistence pairs}) in the persistence diagram \emph{(b)}. In the presence of noise \emph{(c)}, spurious features are contained in the diagram in the vicinity of the diagonal \emph{(d)}. This enables their visual identification and removal via \emph{topological simplification} \emph{(e)-(f)}. In this terrain dataset, the shape of each hill can be nicely captured by the merge tree \emph{(f)}, in particular by considering the regions associated to its leaf arcs \emph{(g)}.}
  \label{fig:theoreticalBackground}
\end{figure*}

This section documents
the technical context of
HVI fast imaging.

\subsection{Impact physics}
\label{subsec_ImpactPhysics}
We briefly recall here the
physical phenomena occurring
when a small projectile impacts a large target at hypervelocity  (\autoref{fig:applicationContext}), since they will be considered later,
when formulating initial hypotheses
(\autoref{sec_domainInfo}).

\noindent
\textbf{\emph{(i)} Phenomenology:}
\david{If the impact velocity is high enough (\emph{i.e.,} typically higher than a few km/s), both materials undergo a shock compression \cite{kinslow1970high}. In the vicinity of the impact point, large deformations occur. In the case of brittle targets (such as rocks or ceramics), these deformations are the result of crack formation and propagation,
leading to many separate debris fragments \cite{Collins2004}. Moreover, when the compression is realeased, significant velocity can be imparted to the fragments, resulting in material (or debris) ejection \cite{melosh85}. During its propagation in the target, the shock amplitude decreases. The balance between the initial shock pressure, its attenuation during propagation,
the
strength and the microstructure of the target,
yields
a finite-size crater, whose
dimension
and shape
result from
this complex and transient process \cite{zukas1982impact}.}

\noindent
\textbf{\emph{(ii)} Debris ejection processes:}
\david{Given the
transient nature of
cratering,
simulations have allowed
a few
insights in the
mechanisms leading to debris ejection.
\emph{Jetting} \cite{Kurosawa2015}
\julien{is a high-pressure, early-time phenomenon caused by extreme pressure gradients at impact, leading to the ejection of very high-velocity, fine debris.
\emph{Spallation} \cite{Kurosawa2018}
results from the shockwave,
causing the detachment of larger, slower fragments.}
Below the projectile impact area, the fragmentation may be more likely due to shear stress under confining pressure \cite{Collins2004}.}

\noindent
\textbf{\emph{(iii)} Analogy with lasers:}
\david{As discussed above,
debris
ejection
is
mostly driven by the shock and release waves generated by the HVI,
and it is not only due to the projectile.
Thus, it is also possible to investigate cratering
with
high intensity lasers (\autoref{sec_caseStudies}), that allow similar dynamical loading  \cite{AUBERT2025}, along with easier diagnostic implementation \cite{SEISSON2016}.}



\subsection{\david{Ejecta video recording}}
\label{sec_experimentalData}
HVI monitoring
typically involves the recording by a high-frame-rate camera of the \david{impact} of the projectile against the target under study \david{and the subsequent debris ejection} (\autoref{fig:applicationContext}). The projectile (\david{here} an \todo{aluminum}
\julien{ball}
of \theophane{2} mm radius) simulates an orbital debris hitting the shielding system of a satellite, represented in the experiment by a \david{porous and isotropic} graphite
target.
In our setup, the
projectile is launched at
\theophane{4.79 km/s} by a two-stage light-gas gun (2SLGG HERMES), a device
capable of accelerating a millimeter size projectile at velocity ranging from 1 to 10 km/s.
\david{The impact} is captured by a ultra-high-speed camera, \todo{at a typical} frame-rate of \theophane{200,000} images per second. 
(\autoref{fig:applicationContext}).
\david{As discussed in \autoref{subsec_ImpactPhysics}, debris ejection can also be generated with
laser-based HVI.
Our database also
includes acquisitions obtained with this modality, with
shock waves induced by
a high-power pulsed laser, and filmed by a specific, intensified, multi-channel ultra-high-speed camera, at a typical frame-rate of $2$ million images per second.
}

Specifically, our database contains
\theophane{7} acquisitions of projectile launches, including \theophane{5} and \theophane{2} with respective impact angles of 90 and 45°.
These consist in 2D grayscale videos, counting \theophane{from 400 to 1,050} frames, with a resolution \theophane{from $384 \times 176$ to $384 \times 256$} pixels.
Our database also contains \theophane{an} acquisition of impact generated via laser pulse (90° impact angle), for a resolution of \theophane{$400 \times 250$} over \theophane{256} frames.

For each experiment, 
\revision{each video frame undergoes a basic pre-smoothing step based on 
1-neighborhood pixel averaging prior to file export. Moreover,}
a depth scan of the target is conducted after impact, to acquire the crater's 
shape, often analyzed
via
sectional profiles (\autoref{fig:applicationContext}). This
also enables the measure of the ejected mass.





\subsection{Observations, assumptions and hypotheses}
\label{sec_domainInfo}
The visualization of the HVI acquisitions by the domain experts triggered several observations (regarding the acquisition process itself), accompanied by specific physical assumptions, as well as initial hypotheses regarding the behavior of the debris.
The observations and assumptions
will be
exploited   to incorporate domain knowledge
(\autoref{sec_topoTracking})
and physical priors
(\autoref{sec_physics})
into our tracking approach,
while our case study will evaluate the corroboration of the initial hypotheses
(\autoref{sec_caseStudies}).

\noindent
\textbf{Observation O1: Important vignetting.}
Given the high-frame-rate camera setups,
hypervelocity impact images are typically affected by strong vignetting artifacts, resulting in large variations in pixel intensity and contrast throughout the image. As shown in \autoref{fig:dataChallenge}, this challenges methods based on intensity thresholding for debris \emph{characterization}.

\noindent
\textbf{Observation O2: Debris shape and intensity variation.}
Important shape and size variations can be observed among the debris population.
Also,
since each debris
fragment
travels in 3D with a spin motion, its 2D projection on the image yields a shape that varies through time.
This, in combination with
vignetting artifacts,
generates debris fragments
that, taken \emph{individually}, vary
in shape, size and pixel intensity over time, sometimes drastically.
This challenges debris \emph{characterization}.

\noindent
\textbf{Observation O3: Insufficient capture frequency.} Even at very high frame-rates, a single debris
fragment
can travel so fast that its projections in consecutive frames do not overlap (\autoref{fig:dataChallenge}).
This challenges methods based on object overlap estimation for debris \emph{tracking}.
%

\noindent
\textbf{Observation O4: Important occlusion.} Since the acquisition captures with a 2D video a 3D phenomenon, important occlusions inevitably occur. This effect is particularly important
at
the beginning of the
recording,
when the majority of the debris are captured in the form of a large blob, that progressively spreads into individual pieces. Occlusion is also clearly noticeable later in the ejection, when multiple debris may occlude each other. While these  objects overlap in the 2D image, they should still be characterized as distinct debris fragments. This also challenges
methods based on
overlap estimation for debris \emph{tracking}.

In addition to these observations, several physical assumptions can be made given the order of velocities captured by these experiments.

\noindent
\textbf{Physical assumption PA1: Constant individual velocity.} Given the
\david{recording duration (< 10 ms) and the debris velocities (> 1 m/s), the effect of gravity can be neglected.} Then,
each debris
fragment
can be safely assumed to travel according to an \emph{individual} velocity vector that is constant through the acquisition.


\noindent
\textbf{Physical assumption PA2: Constant individual mass.} Given
the debris velocities,
%
it can be safely assumed that there is no significant
interaction
among debris.
It follows that
debris fragments
do not split into multiple pieces after the initial impact.

\noindent
\textbf{Physical assumption PA3: Uniform density.} Given
the target material, all the debris can be considered to have the same volume density.

After
observing
the acquisitions, the domain experts formulated several hypotheses that they would like to investigate
visually.


\noindent
\textbf{Hypothesis H1: \fv{Multi-phase} 
ejection.}
\fv{The debris ejection seems to occur in several phases. For
projectile-based HVI,
it is assumed that these different phases correspond to the different fragmentation processes discussed in \autoref{subsec_ImpactPhysics} (e.g., \emph{jetting} and \emph{spallation}).
}

\noindent
\textbf{Hypothesis H2: Velocity disparity \fv{correlated to angle trajectory.}} \fv{According to previous studies \cite{HEBERT2022, Reynier, POLYCARBONATE},
a signature of the different processes
(e.g., \emph{jetting} and \emph{spallation}, \autoref{subsec_ImpactPhysics})
can be found in debris statistics, especially angle-velocity and mass-velocity
distributions.
  }

Our work provides a reliable framework specifically tailored for debris tracking, supporting
the computation of statistical summaries that enable the visual investigation of the above hypotheses (\autoref{sec_caseStudies}).


%% file: background.tex
\section{Background}

This section presents the
technical background to our work. We refer
to
textbooks \cite{edelsbrunner09} for an introduction to computational topology.

\subsection{Input representation}
\label{sec_inputData}
Each
input
acquisition is given
as
a
\revision{\emph{floating-point}}
time-varing scalar field $F = \{f_{1}, f_{2}, \dots, f_{\frameNumber}\}$, where
$\frameNumber$ is the number of time steps. Each time step is modeled as a piecewise linear (PL) scalar field
$f_{t} : \domain \rightarrow \mathbb{R}$,
with $t \in \{1, 2, \dots, \frameNumber\}$, which encodes, for convenience, the \emph{opposite} of the pixel intensity \revision{$I_t$ at time $t$}, on the triangulation $\domain$ of the input 2D pixel grid\revision{, i.e., $\forall v \in \domain, f_t(v) = -I_t(v)$}.
For brevity, we will omit the notation
$_t$
when
non-ambiguously
focusing on a single time step.

For a given time step,
$f$
values are provided on the vertices of $\domain$ and $f$ is interpolated with barycentric coordinates on the other simplices of $\domain$.
$f$ is enforced to be injective on the vertices with a variant of simulation of simplicity \cite{edelsbrunner90}.
In practice, we represent each time step as an elevated 3D terrain (\autoref{fig:theoreticalBackground}), where the elevation coordinate (i.e., $Z$) is given by the $f$ value of each vertex.


The sub-level set of $f$ for the isovalue $i$, noted $\sublevelset{f}(i)$, is defined as the subset of $\domain$ valued below the isovalue $i$.
The super-level set, noted $\superlevelset{f}(i)$, is defined symmetrically (subset of $\domain$ above $i$).
As $i$ continuously evolves from $-\infty$ to $+\infty$, the \emph{topology} of 
$\sublevelset{f}(i)$ changes at precise locations, called \emph{critical points} 
\cite{edelsbrunner09}. New connected components of $\sublevelset{f}(i)$ emerge 
on minima of $f$. On the saddle points of $f$, either two connected components 
of $\sublevelset{f}(i)$ merge, or
a single component loops back onto itself to form
a topological handle in $\sublevelset{f}(i)$ (saddles can be enforced to be non-degenerate via saddle unfolding \cite{edelsbrunner09}). Finally, topological handles are completely filled on maxima. As shown in \autoref{fig:theoreticalBackground},
the critical points
capture the topological features
of the scalar field: minima, saddles and maxima encode pits, valleys and peaks.

\subsection{Persistence diagrams}
\label{sec_persistenceDiagrams}
Real-life data is often affected by noise. This is particularly the case with our hypervelocity impact acquisitions. As shown in \autoref{fig:theoreticalBackground}, in the presence of noise, the slightest oscillation in data values yields spurious critical points. Topological persistence \cite{edelsbrunner02, edelsbrunner09} addresses this issue. It is an established framework for characterizing the importance of critical points. In our setup, it will be instrumental for the detection of the most salient critical points, irrespective of vignetting effects (brightness and contrast variations throughout the image).

In the domain, each topological feature (i.e., connected component, handle) of $\sublevelset{f}(i)$ can be associated with a unique pair of critical points, called a \emph{persistence pair} $(c, c')$, corresponding to its \emph{birth} and \emph{death}. The Elder rule \cite{edelsbrunner09} provides a mechanism for arranging the set of critical points into persistence pairs.
For instance, if two connected components of $\sublevelset{f}(i)$ merge at a saddle point $c'$, the
\emph{younger} component (created last, in $c$) \emph{dies}, in favor of the \emph{older} one (created first). In practice, the life-span of connected components can be tracked with a Union-Find data structure \cite{cormen}.
In 2D, handle creation (on saddles) and destruction (on maxima) can also be tracked
with a Union-Find data structure by duality \cite{edelsbrunner02, guillou_tvcg23}, by considering the
super-level sets of $f$.

Persistence pairs can be represented visually in the persistence diagram, noted $\diagram(f)$. Specifically, each persistence pair $(c, c')$ is embedded as a vertical bar in 2D, such that the $x$-coordinate of the bar is given by $f(c)$
and that the $y$-coordinates of its bottom and top are given by $f(c)$ and $f(c')$ respectively. Then, the life-span of the corresponding feature can be directly read from the diagram as the length of the bar, $|f(c') - f(c)|$, called \emph{topological persistence}
\cite{edelsbrunner02}.
As shown in \autoref{fig:theoreticalBackground}, the most prominent features in the data are captured with large bars in the diagram, which can be easily distinguished from the noise, characterized by short bars near the diagonal.
In contrast to isovalue thresholding approaches which are challenged by vignetting effects (\autoref{fig:dataChallenge}), the salience evaluation provided by topological persistence (based on the capture of large intensity variations) overcomes this issue and it will be instrumental in our framework to robustly identify debris.

Persistence diagrams are often simplified, by discarding the bars shorter than the expected noise level. Topological simplification
\cite{EdelsbrunnerMP06,
Lukasczyk_vis20}
is a process that enables the estimation of a new function $\widehat{f}$, which is close to the input function $f$, and whose topology is exactly described by the simplified diagram. Topological simplification is an established pre-processing step in practice, as it enables the access  to simplified versions of the input, for multi-scale topological analysis.

\subsection{Merge trees}
\label{sec_mergeTrees}
As described next,
debris are characterized and tracked in our work via the critical points of the time-varying scalar field.
Once a debris is identified and tracked, its geometry needs to be evaluated in order to estimate its mass (\autoref{sec_geometry}). For that, we rely on the \emph{merge tree} \cite{
carr00, gueunet_tpds19} to extract the geometry of the hills captured by the maxima (\autoref{fig:theoreticalBackground}).

The merge tree of $f$ is a 1-dimensional simplicial complex defined as the quotient space $\mergetree(f) = \domain / \sim$ by the equivalence relation $\sim$, which states that two points $p_1$ and $p_2$ are equivalent if $f(p_1) = f(p_2)$ and if $p_1$ and $p_2$ belong to the same connected component of super-level set $\superlevelset{f}\big(f(p_1)\big)$. In short, this tree tracks the creation of connected components of $\superlevelset{f}(i)$ at its leaves, and merge events at its interior nodes (\autoref{fig:theoreticalBackground}). This tree can also be computed by maintaining a Union-Find
while sweeping the data from top to bottom values \cite{carr00, gueunet_tpds19}. Interestingly, the construction of these trees maintains the relationship that assigns each vertex of $\domain$ to its corresponding arc in the tree. Then, after computation, $\domain$ can be precisely partitioned into the regions corresponding to its arcs. In particular, the regions corresponding to its leaf arcs nicely capture the hills associated to each maximum (\autoref{fig:theoreticalBackground}).

%% file: tracking.tex
\section{Topology tracking}
\label{sec_topoTracking}

This section describes how to extend an off-the-shelf topology tracking framework \cite{soler_ldav18}, to incorporate \emph{domain knowledge} about the acquisition, in particular regarding the \emph{observations} documented in \autoref{sec_domainInfo}.

\subsection{Debris characterization}
\label{sec_characterization}
This section describes the characterization of debris in our work. While the pipeline described in this section is typical of the state of the art \cite{edelsbrunner09}, we emphasize here details that are specific to our application context (\autoref{sec_applicationContext}) and which motivate the design of our framework.

\noindent
\textbf{\emph{(i)} Feature specification.}
As described in
\autoref{sec_inputData}, the scalar field $f$ considered at each time step is the \emph{opposite} of pixel intensity values. Moreover, each time step is represented by an elevated terrain, where the elevation coordinate (i.e., $Z$) is given by $f$. Then,
in this framework, debris are characterized by prominent hills in the terrain.

\begin{figure}
  \centering
  \includegraphics[width=\linewidth]{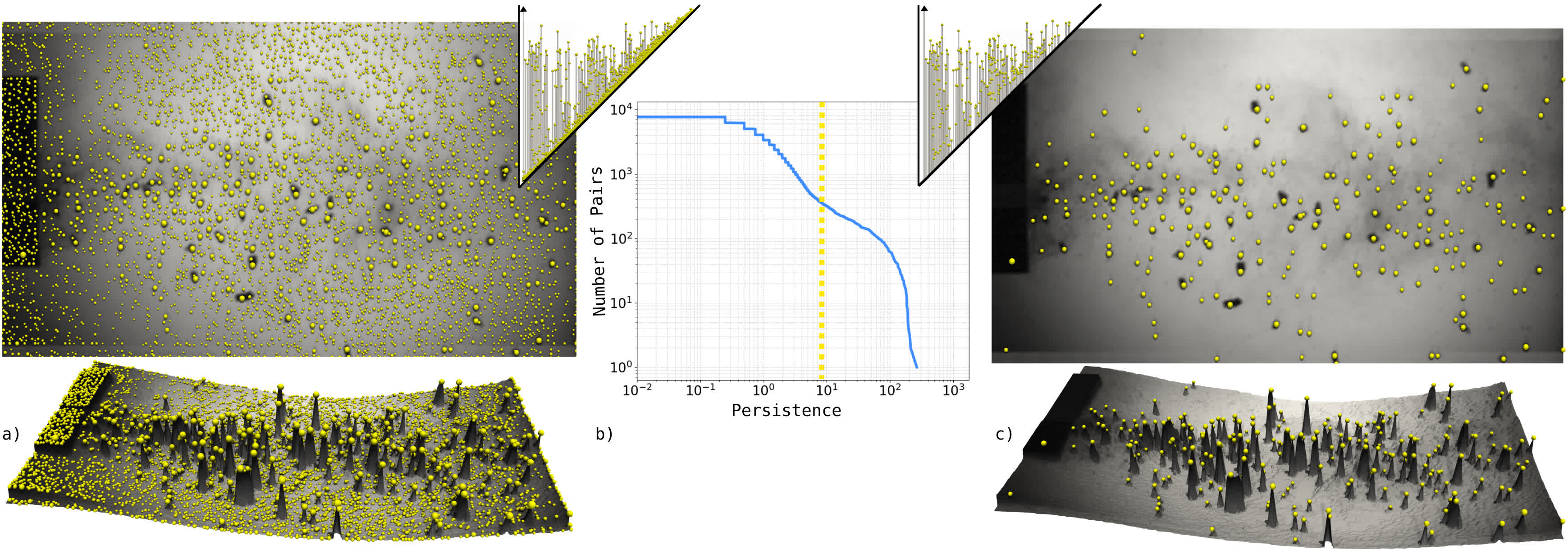}
  \caption{Debris characterization by persistence thresholding: each time step is represented as a 3D terrain. Noise and vignetting artifacts result in the presence of many spurious features, for diverse intensity values  \emph{(a)}. Maxima, in yellow, are represented with spheres  whose radius is a function of \emph{topological persistence}. The persistence curve
  \emph{(b)}
  leads to the
  visual
  identification (inflection point) of a persistence threshold separating noise from features. This enables the characterization of the debris as the most persistent maxima in this terrain representation \emph{(c)}.}
  \label{fig:debrisCharacterization}
\end{figure}

\noindent
\textbf{\emph{(ii)} Topological characterization.} Peaks in the elevated terrains of $f$ are captured by the \emph{maxima} of $f$. However,
as illustrated in \autoref{fig:debrisCharacterization}, the acquired images are subject to noise, leading to the identification of many spurious maxima associated with persistence pairs of low persistence (\autoref{sec_persistenceDiagrams}). Thus, we only consider in the following the maxima involved in persistence pairs with a persistence larger than a fraction $\persistenceThreshold$ of the overall function range, \autoref{fig:debrisCharacterization}\emph{(c)}. This fraction can be visually identified at an inflection point, \autoref{fig:debrisCharacterization}\emph{(b)}, in the persistence curve (complementary cumulative function of persistence pairs). As described in \autoref{sec_results}, this parameter is set once for all
(see \autoref{sec_parameterSetting} for a description of our parameter setup 
protocol). After this initial persistence thresholding, 
\emph{topological simplification} (\autoref{sec_persistenceDiagrams}) 
\revision{will be performed}
such that 
our subsequent analysis (\autoref{sec_geometry}) also discards spurious maxima 
consistently. Note that, since \emph{topological persistence} is defined as a 
difference between the function values of paired critical points 
(\autoref{sec_persistenceDiagrams}), it
naturally addresses
the vignetting artifacts (\emph{observation O1} from \autoref{sec_domainInfo}). Indeed,
while pixel intensity progressively decreases away from the center of the 
image, 
\revision{debris persistence}
remains nearly constant over the domain, easing 
their identification.


%
%
%

\subsection{Debris tracking}
\label{sec_tracking}
Soler et al. \cite{soler_ldav18} presented a feature tracking approach based on the optimal assignment of the extrema between two consecutive time-steps $f_i$ and $f_j$ of a time-varying scalar field. In our work, debris can be followed by tracking
persistent maxima
over time. Thus, we focus on this off-the-shelf tracking approach
and we document here
adjustments to make it fit  our application context (\autoref{sec_applicationContext}).

\noindent
\textbf{Off-the-shelf tracking framework.} Here we recap the approach by Soler et al. \cite{soler_ldav18}, in a way that eases the formalization of
the modifications we considered in our work.
First, each input saddle-maximum persistence pair $p = (c, c')$ of
a diagram $\diagram(f)$
is represented as a point in $\mathbb{R}^5$, where the first three coordinates of $p$ are the 3D coordinates of the maximum $c'$ in the domain and the remaining two coordinates are the \emph{birth} and \emph{death} values of $p$, i.e., $f(c)$ and $f(c')$ respectively. Second, to account for a domain-dependent disparity between 3D distances and function values, diagrams undergo a \emph{scaling procedure}. Specifically,
each resulting $5$-dimensional point $p$ is scaled,
with the element-wise multiplication of its coordinates by those of a
\emph{scaling vector} $s \in \mathbb{R}^5$.

Third, in order to
match the saddle-maximum pairs of a first diagram $\diagram(f_i)$ to a second $\diagram(f_j)$, both diagrams undergo an \emph{augmentation procedure}. This process aims at enabling the modeling of feature creation or destruction between time steps. For that, for each saddle-maximum pair $p_i \in \diagram(f_i)$, a corresponding \emph{dummy pair}  $\projection p_i$ is created. It is also modeled as a $5$-dimensional point, whose first three coordinates correspond to the 3D coordinates of the mid-point between the critical points $c$ and $c'$, and whose last two coordinates, modeling birth and death, are given identical values: $\big(f(c) + f(c')\big)/2$.
Then, the diagrams $\diagram(f_i)$ and $\diagram(f_j)$ are augmented with all the \emph{dummy pairs} of the other:
\begin{eqnarray}
  \nonumber
 \diagram'(f_i) = \diagram(f_i) \cup \{
\projection(p_j) ~ | ~ p_j \in \diagram(f_j)
\}\\
\nonumber
  \diagram'(f_j) = \diagram(f_j) \cup \{
\projection(p_i) ~ | ~ p_i \in \diagram(f_i)
\}.
\end{eqnarray}
In short, this augmentation procedure inserts dummy features with zero persistence allowing a pair $p_i$ to either be matched with a pair in $\diagram(f_j)$ or to its dummy pair $\Delta p_i$. After augmentation, both diagrams have the same size by construction (i.e., $|\diagram'(f_i)| = |\diagram'(f_j)|$).

The weight $\weight(p_i, p_j)$ associated to the matching a persistence pair $p_i \in \diagram'(f_i)$ to another one
$p_j \in \diagram'(f_j)$ is then given by
the $L_2$-norm between the corresponding $5$-dimensional vectors:
\begin{eqnarray}
  \nonumber
\weight(p_i, p_j)
= \| p_i - p_j
\|_2.
\end{eqnarray}
In the special case where both $p_i$ and $p_j$ are \emph{dummy pairs}, $\weight(p_i, p_j)$ is set to zero. Then, the
optimal
matching $\phi^*$ between the maxima of $f_i$ and those of $f_j$ is a bijection from $\diagram'(f_i)$ to $\diagram'(f_j)$, computed as the minimizer of the following assignment functional $E(\phi)$:
\begin{eqnarray}
\nonumber
E(\phi) = \sum_{p_i \in \diagram'(f_i)} \weight\big(p_i, \phi(p_i)\big).
\end{eqnarray}
In short, $\phi^*$ is a matching which minimizes the sum of the costs of mapping the maxima of $f_i$ to those of $f_j$ according to the weight $\weight$.

\noindent
\textbf{Modification M1: Pre-rotation.}
In our application, debris mostly travel along a direction that is
\todo{orthogonal to the target surface} (\autoref{sec_applicationContext}). Thus, in case of
oblique
projectile launches (e.g., 45°
angle), we first rotate the data in the $XY$ plane, such that the $X$-axis becomes orthogonal to the target surface (and thus aligned with the main direction of
travel).

\noindent
\textbf{Modification M2: Pre-scaling.} Since each time step is represented as a 3D terrain (\autoref{sec_inputData}), the $Z$-coordinate of each maximum already captures its scalar value. Thus, we consider the following \emph{scaling vector}
$\scalingVector = (\scalingVector_X, \scalingVector_Y, \scalingVector_Z, 0, 0)$ (i.e., with $0$ scaling for birth and death values, the death value being already encoded along the $Z$-coordinate). With this
modification,
the optimization of the matching between 
maxima
(described above)
becomes a simple assignment problem between point clouds in 3D, the simplest form of optimal transport \cite{PeyreC19}. In particular,
in order to mitigate the insufficient capture frequency of the employed cameras (\emph{observation O3}, \autoref{sec_domainInfo}), we will typically use of a low value for $\scalingVector_X$ (as documented in \autoref{sec_parameterSetting}), since debris can undergo large displacements along this direction from one time step to the next.

\noindent
\textbf{Modification M3: Maximum appearance and disappearance.} At the beginning of the acquisition, due to important occlusion (\emph{observation O4}, \autoref{sec_domainInfo}), the majority of the debris are initially captured by a large blob, which progressively spreads into individual pieces, leading to the appearance of more and more persistent maxima with time. Symmetrically, towards the end of the acquisition, debris progressively exit the image, leading to a decrease in the number of persistent maxima. Therefore, for these two reasons, maximum appearance and disappearance still need to be modeled in our assignment problem. Originally, given a saddle-maximum pair $p_i = (c, c')$,
the selected tracking framework \cite{soler_ldav18} considered a 3D embedding for its dummy pair $\projection p_i$ at the midpoint between the critical points $c$ and $c'$. However, in our applications, saddles tend to be located in the background of the image, which tends to be a flat plateau. As a consequence, the saddle $c$ associated to a maximum $c'$ can be arbitrarily far from it, leading to arbitrary destruction weights in our setup. To address this, we simply re-define the weight for matching a persistence pair $p_i$ to its dummy pair $\projection p_i$ to a constant value: $\weight(p_i, \projection p_i) = \destructionCost$.
\todo{This parameter is stable across multiple acquisitions and thus, we set it once for all
to a conservative value
(see \autoref{sec_parameterSetting} for our parameter setup
protocol).}
This means that if no pair $p_j \in \diagram'(f_j)$ is located within a ball of radius $2\destructionCost$ around $p_i$, $p_i$ will have to be considered as destroyed from time step $i$ to $j$. This
accounts
for debris leaving the image. Symmetrically, if no pair $p_i \in \diagram'(f_i)$ is located within a ball of radius $2\destructionCost$ around $p_j$, this means that the debris captured by $p_j$ just emerged a time step $j$ as an individual piece, e.g., from the initial debris blob
(\emph{observation O4}, \autoref{sec_domainInfo}).

%% file: physics.tex
\section{Physics-driven post-processing}
\label{sec_physics}

This section describes how to extend the tracking approach presented
above,
in order
to incorporate the
\emph{physical assumptions} provided
by
domain experts (documented in \autoref{sec_domainInfo}). Specifically,
it
presents how to post-process the raw trajectories produced by the initial tracking into physically plausible trajectories, that can be directly exploited
for the reliable analysis of debris ejection in hypervelocity impacts.


\subsection{From trajectory paths to trajectory segments}
\label{sec_physicalTrajectories}
\begin{figure}
  \centering
  \includegraphics[width=\linewidth]{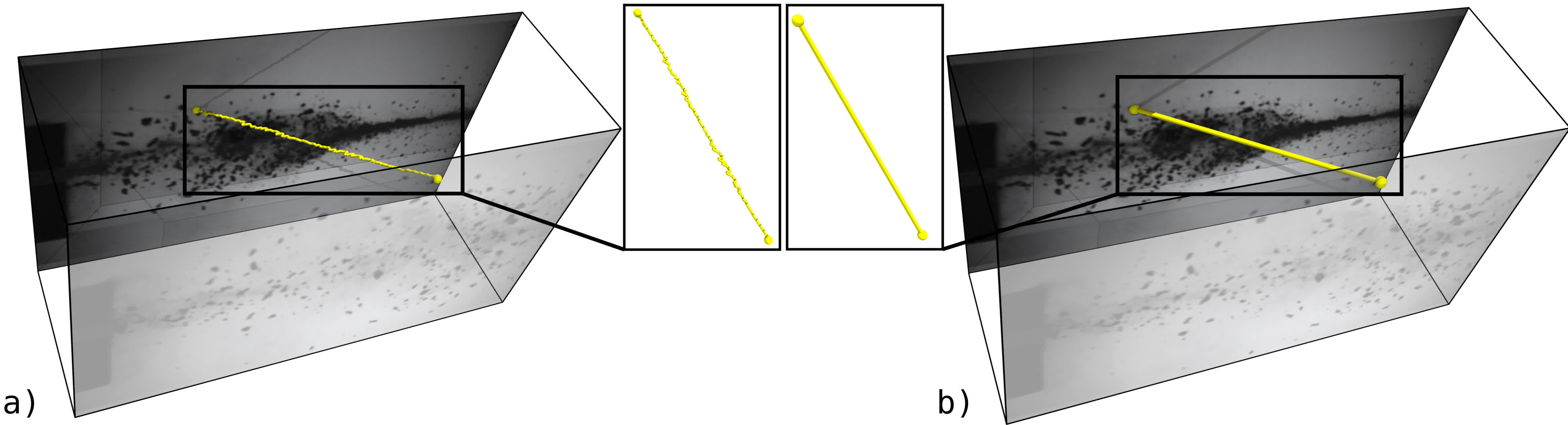}
  \caption{Each trajectory \emph{path} computed by the topology tracking \emph{(a)} is transformed into a trajectory \emph{segment} \emph{(b)} via linear regression, given the \emph{physical assumption PA1} of
  per-fragment constant velocity vector.}
  \label{fig:constantSpeedHypothesis}
\end{figure}
The debris tracking algorithm described in the previous section produces a set $\trajectories$ of
$\trajectoryNumber$
\emph{trajectory paths} $\trajectories = \{ \trajectory_1, \trajectory_2, \dots
\trajectory_{\trajectoryNumber}\}$,
each path $\trajectory_i$ being 
an ordered set of $n_{\trajectory_i}$ points, representing the 
maxima
matched between consecutive time steps: $\trajectory_i = \{ \criticalPoint^i_1, \criticalPoint^i_2, \dots, \criticalPoint^i_{n_{\trajectory_i}}\}$, each 
maximum
$\criticalPoint^i_k$ being associated with its  time step through the map $t(\criticalPoint^i_k)$.

As documented in \autoref{sec_domainInfo}, taken individually, each debris
fragment
can be assumed to travel along a constant velocity vector (\emph{physical assumption PA1}).
Thus,
each debris trajectory should eventually be modeled with a unique
line segment (\autoref{fig:constantSpeedHypothesis}).
For that, we 
consider for 
each trajectory $\trajectory_i$ a unique 2D line $\velocityLine_i(t) = a_i t + b_i$, where $b_i$ is the debris $(X, Y)$ position extrapolated at temporal origin and where $a_i$ is the estimated 2D \emph{velocity vector} of the corresponding debris. Specifically, the constants $a_i$ and $b_i$ are estimated via standard least squares linear regression \cite{SeberLee2003, Kalman1960}. Also, for each line $\velocityLine_i$, we store the
\emph{trajectory segment}
$\trackingSegment_i = [\velocityLine_i(t_s^i), \velocityLine_i(t_e^i)]$
over which the debris was actually followed by the trajectory path $\trajectory_i$, i.e., $t_s^i = t(\criticalPoint_1^i)$ and $t_e^i = t (\criticalPoint_{n_{\trajectory_i}}^i)$. Finally, we maintain a map
between each 
\emph{trajectory
segment} $\trackingSegment_i$ and its initial \emph{trajectory path} $\trajectory_i$.



%


\begin{figure}
  \centering
  \includegraphics[width=\linewidth]{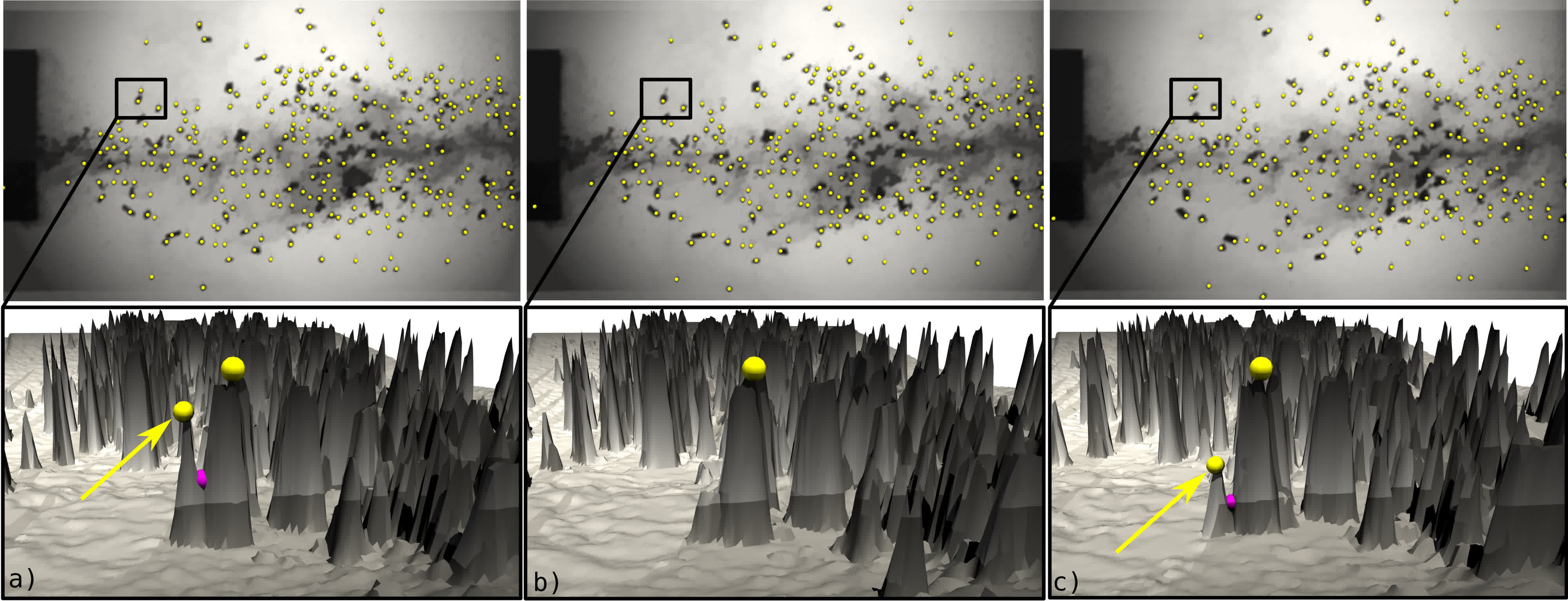}
  \caption{Gaps in trajectory paths can occur when 
  a
  debris
  fragment, identified by a persistent maximum
  (arrow) at time step $i$ \emph{(a)}, is associated at the next time step $j$ with a maximum
  below the persistence cutoff \emph{(b)}. At a later time step $k$ \emph{(c)}, the corresponding maximum can become persistent again, yielding a tracking gap between the time steps $i$ and $k$.}
  \label{fig:featureCreationDestruction}
\end{figure}

\subsection{From trajectory segments to trajectory lines}
\label{sec_trajectoryContinuation}
\begin{figure}
  \centering
  \includegraphics[width=\linewidth]{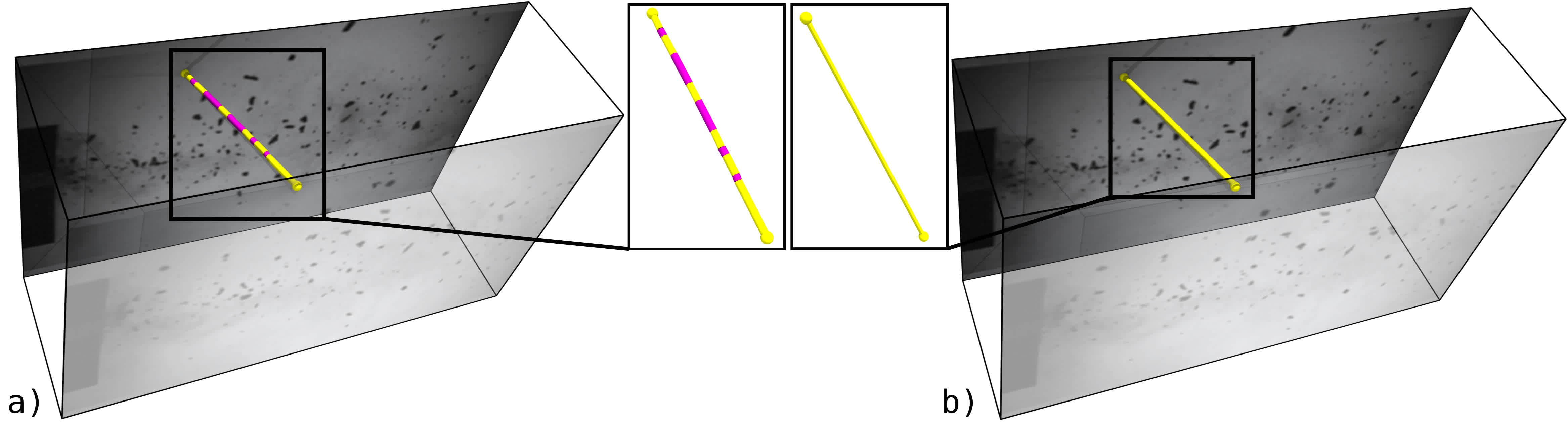}
  \caption{Consecutive \emph{trajectory segments} \emph{(a)}, in yellow, are concatenated into a single  \emph{trajectory line} \emph{(b)} under the \emph{physical assumption PA2} of
  per-fragment
  constant mass,
  by condering
  space-time criteria (\autoref{sec_trajectoryContinuation}).}
  \label{fig:materialPreservation}
\end{figure}
As documented in \autoref{sec_domainInfo}, there is no significant secondary impact among debris and, consequently, debris do not merge or split (\emph{physical assumption PA2}).
However, the practical setup of the acquisitions challenges this hypothesis. As reported in \emph{observation O4} (\autoref{sec_domainInfo}), since the acquisition captures in 2D a phenomenon that is intrinsically 3D, important occlusion inevitably occurs. Specifically, two debris, characterized by two independent 
maxima 
at time step $i$ can 
\emph{overlap} in 2D at the following time step $j$, resulting in a single,
persistent maximum.
In that scenario, the \emph{trajectory segment} (\autoref{sec_physicalTrajectories}) of one of the two debris will artificially terminate at time step $j$. Moreover, this debris
may be tracked again by a new, independent segment at a later time step $k$, resulting overall in a \emph{gap} in its tracking (between time steps $j$ and $k$).
Furthermore, as reported in \emph{observation O2} (\autoref{sec_domainInfo}), the shape and intensity of each debris can vary significantly through time. Consequently,
a debris characterized by a maximum that is persistent enough at time step $i$ may be characterized at time step $j$ by a maximum whose persistence falls below the cutoff threshold (\autoref{sec_characterization}). In that case, the trajectory segment will also artificially terminate at step $j$. Then, the debris may be tracked again by a new, independent segment at a later time step $k$, as soon as the associated maximum becomes again persistent enough,
also resulting in \emph{gap} in the tracking (between time steps $j$ and $k$, \autoref{fig:featureCreationDestruction}). In both
cases, the gaps in the tracking will result in an \emph{over-estimation} of the number of trajectories, since a given debris may be tracked by multiple, consecutive \emph{trajectory segments}.

To address
this,
a trajectory continuation procedure is introduced, to concatenate the \emph{trajectory segments}
corresponding to a single
fragment
(\autoref{fig:materialPreservation}).
To
evaluate if two consecutive segments
$\trackingSegment_i$ and
$\trackingSegment_j$ correspond to the same debris, we consider three complementary criteria.

\noindent
\textbf{\emph{(i)} Temporal proximity.} The temporal gap  between the two segments
$\temporalGap_{ij} = t_s^j - t_e^i$
must be in the interval $[0, \temporalGap_{max}]$.

\noindent
\textbf{\emph{(ii)} Directional consistency.} The absolute value of the angle $\myangle_{ij}$ between the
velocity
vectors $a_i$ and $a_j$ must be
in the interval $[0, \myangle_{max}]$.

\noindent
\textbf{\emph{(iii)} Spatial continuity.} The
fragment
position extrapolated from $\trackingSegment_i$ at time $t_s^j$ must coincide with the start of $\trackingSegment_j$:
the distance $d_{ij} = ||\velocityLine_i(t_s^j) - \velocityLine_j(t_s^j)||_2$ must be smaller than a threshold $d_{max}$ (close to $0$).

Then, given a segment $\trackingSegment_i$, we first evaluate its 
available
candidates for concatenation based on the above thresholds ($\temporalGap_{max}$, $\myangle_{max}$, $d_{max}$). Then, $\trackingSegment_i$ is greedily assigned to the candidate segment $\trackingSegment_j$ which minimizes the sum of the above penalties.
At this point, each debris is now represented by
an ordered set of $n_{i}$ trajectory segments $\segmentSequence_i = \{\trackingSegment_1^i, \trackingSegment_2^i, \dots, \trackingSegment_{n_i}^i\}$, for which we estimate a unique \emph{trajectory line} $\trajectoryLine_i$, also via
least squares
linear regression (\autoref{sec_physicalTrajectories}).
Then, the output of this stage is a collection  $\lineCollection$ of $n$ \emph{trajectory lines} $\lineCollection = \{\trajectoryLine_1, \trajectoryLine_2, \dots, \trajectoryLine_n\}$, where we maintain, for each trajectory line $\trajectoryLine_i$, a map to its set of trajectory segments $\segmentSequence_i$.

%
%
%
%
%
%
%


\begin{figure}
 \centering
 \includegraphics[width=\linewidth]{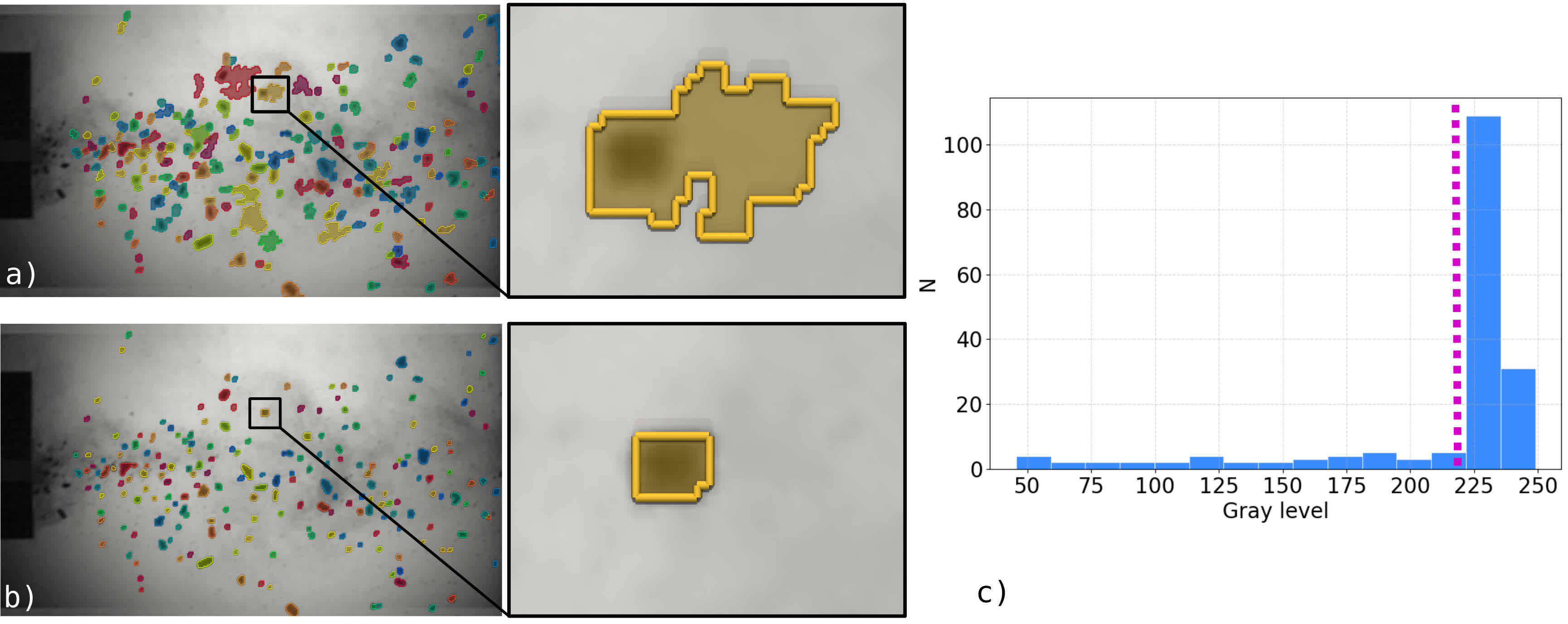}
	\caption{The regions associated to the leaf-arcs of the merge tree provide a first estimation of the debris shapes \emph{(a)}. Since saddles can be located arbitrarily far in the flat background, this segmentation tends to over-estimate debris shape. For each
	of these regions,
	we apply Otsu's algorithm \cite{otsu79} to
	automatically identify an optimal cutoff (dashed line) in its pixel intensity distribution \emph{(c)}, resulting in
	regions
	which individually
	better capture the debris shape, despite vignetting artifacts \emph{(b)}.}
%
%
%
 \label{fig:geometricOtsu}
\end{figure}

\subsection{Physically plausible trajectory lines}
\label{sec_phyicalLines}
At this point, each \emph{trajectory line} $\trajectoryLine_i \in \lineCollection$ represents a single debris
fragment. Given the numerous challenges related to the initial time steps of the acquisition (insufficient capture frequency, large blob representing the entire population of debris, c.f. \autoref{sec_domainInfo}), several trajectory lines of $\lineCollection$ are \emph{outliers} (with few, short, misoriented segments), in particular for the debris identified early in the sequence. Thus, we introduce a last thresholding procedure to remove outliers which are physically aberrant, enabling a focus on the reliable trajectories only.

For that, we consider a geometrical cone of physically plausible trajectories. Specifically, we only maintain in the remainder a trajectory line $\trajectoryLine_i$ if \emph{(i)}
it
intersects the target at $X=0$, \emph{(ii)} does so at a positive time value,  \emph{(iii)} with an angle with the horizontal axis that has an absolute value smaller than
$\coneAngle_{\max}$, and \emph{(iv)} whose X-component of the velocity vector $a_i$ is larger than a lower bound $X_{min}$. The latter constraint on the X-component enables the removal of stationary objects (e.g., the target), while the angle constraint filters out miroriented outliers.
%
%
%
%
This last thresholding produces the final set of trajectory lines $\lineCollection^*$ considered in the rest of our work, and for which the physical validity is guaranteed (plausible cone of trajectories with target intersection).

%
%

\subsection{Debris mass estimation}
\label{sec_geometry}
The tracking approach described above identified a list of physically plausible trajectory lines modeling the displacement of each debris. This first output will enable the reliable estimation of the distribution of debris velocities (by considering the velocity vector $a_i$ of each trajectory line $\trajectoryLine_i$). As described in \autoref{sec_applicationContext}, in addition to this distribution, engineers are also interested in the debris mass distribution.
For this, we rely in this work on a spherical debris shape hypothesis as well as an approximation of the average size of each debris, described below.

For each trajectory \emph{line} $\trajectoryLine_i$, we consider its set of trajectory \emph{segments} $\segmentSequence_i$ (\autoref{sec_trajectoryContinuation}). 
Each trajectory \emph{segment} $\trackingSegment_j \in \segmentSequence_i$ is itself associated to a trajectory path $\trajectory_j$ (\autoref{sec_physicalTrajectories}), modeling the sequence of maxima tracked by the approach described in \autoref{sec_topoTracking}. Given a maximum $m_k^j \in \trajectory_j$, we can retrieve the corresponding time step by considering the map $t(m_k^j)$ (\autoref{sec_physicalTrajectories}).
\revision{In the scalar field $f_{t(m_k^j)}$, let $R$ be the region of the image corresponding to the leaf arc of $\mergetree(f_{t(m_k^j)})$ associated to $m_k^j$,
as discussed in \autoref{sec_mergeTrees}. Then, the shape of the corresponding debris can by nicely captured by the region $R$.}
However,
in certain cases,
since the background of the image tends to be a flat plateau, the saddle associated to $m_k^j$ in the merge tree can be located arbitrarily far in the background, resulting in an over-estimation of the debris size by the region $R$ (\autoref{fig:geometricOtsu}). To address this issue, we rely on the Otsu segmentation algorithm \cite{otsu79} which we restrict to
$R$.
Specifically, this statistical approach optimizes a cutoff value
for $f_{t(m_k^j)}$,
to maximize the variance between the bright and dark pixels, 
%
resulting in an improved approximation of the
shape $\debrisShape_k^j$ of the debris fragment associated to the maximum 
$m_k^j$ (\autoref{fig:geometricOtsu}). 

Next, we evaluate the size of the debris fragment on the image by considering 
the area $\regionArea_k^j$ of $\debrisShape_k^j$. Then, for the trajectory line 
$\trajectoryLine_i$, we consider its average debris area $\debrisArea_i$, by 
averaging the values $\regionArea_k^j$ over 
all the possible values of 
$j$ (segments) and
$k$ (maxima).
Under the hypothesis of a spherical debris shape, the average area 
$\debrisArea_i$ 
enables the estimation of the volume $\debrisVolume_i$ of the debris fragment 
associated to the trajectory line $\trajectoryLine_i$. From this volume 
estimation $\debrisVolume_i$, the mass $\debrisMass_i$ of the debris is finally 
evaluated based on 
the hypothesis of uniform volume density 
(\emph{physical assumption PA3}, \autoref{sec_domainInfo}).

\begin{figure}
  \centering
  \includegraphics[width=\linewidth]{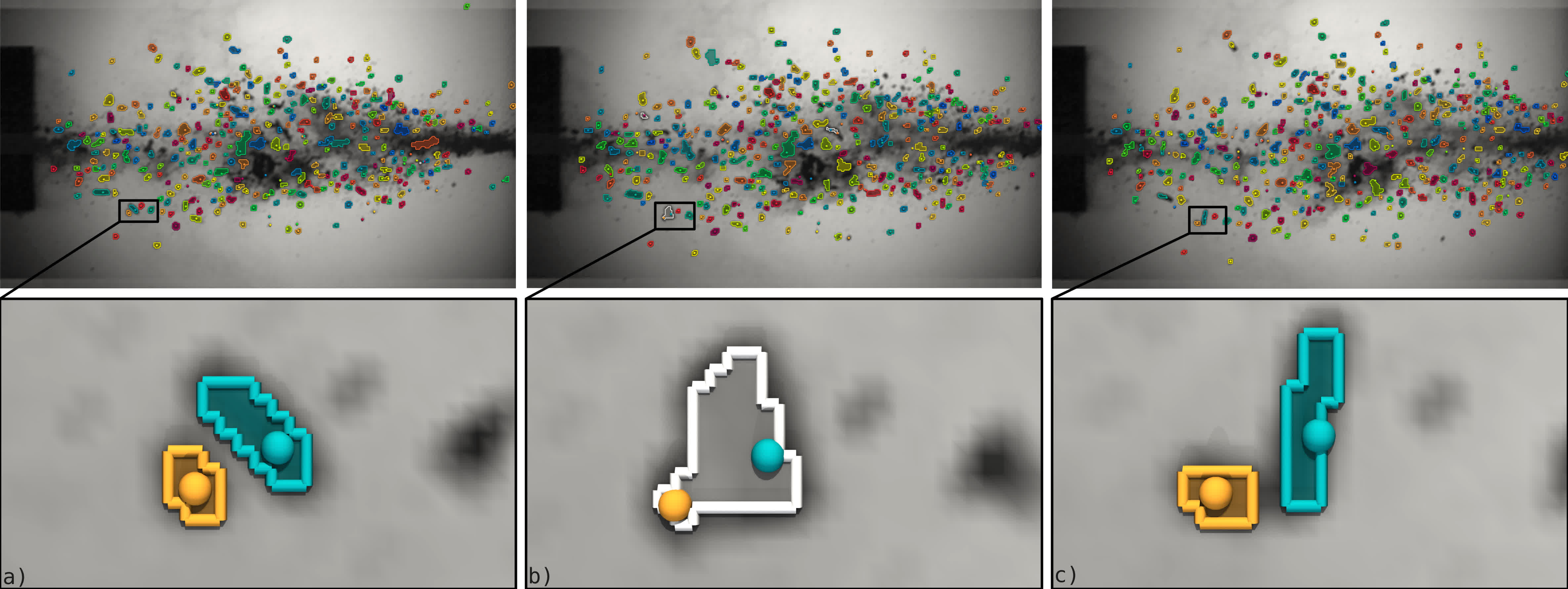}
  \caption{Due to occlusion, independent debris
  fragments
  at frame $i$ \emph{(a)} may momentarily overlap at frame $j$ \emph{(b)}. These configurations are detected in our approach (in white)  by identifying, for each tracking gap (\autoref{sec_geometry}), the merge tree leaf-arc region it
  belongs to
  in the image.}
  \label{fig:geometryOverlap}
\end{figure}

Note that, for a given trajectory line $\trajectoryLine_i$, 
its
set $\segmentSequence_i$ of trajectory segments may not completely cover it,
due to \emph{tracking gaps} derived from occlusion or temporary low persistences
(pink segments in \autoref{fig:materialPreservation}).
For each frame $t$
where a line $\trajectoryLine_i$
has a \emph{tracking gap} (i.e., it does not have any associated segment in $\segmentSequence_i$),
we perform an inclusion test to determine the merge tree leaf arc region $\trajectoryLine_i(t)$ belongs to.
This alternative fallback approach enables
the identification of \emph{overlapping debris} in the image, when a given region is claimed by several trajectory lines, given the above
inclusion
test
(\autoref{fig:geometryOverlap}). When the inclusion test does not identify any leaf-arc region, we simply represent the trajectory line $\trajectoryLine_i$
with an isolated, colored sphere
on the corresponding frame.

%
%

%% file: results.tex
\section{Results}
\label{sec_results}

This section presents experimental results obtained with our approach, implemented in C++ within TTK \cite{ttk17, ttk19}. \todo{Our implementation and our experimental datasets (\autoref{sec_experimentalData}) are available at this address:
\href{https://github.com/tloloum/DebrisTracer}{https://github.com/tloloum/DebrisTracer}.}
\autoref{sec_parameterSetting} documents the parameters of our approach
\revision{and}
the protocol we followed to adjust them, once for all, based on our reference dataset (\autoref{fig:teaser}). It also documents our parameter setting protocol for \emph{TrackMate} \cite{tinevez2017trackmate, Ershov2022}, the established tool used by domain experts, to which we will compare.
\revision{Three domain experts (one junior and two senior scientists, all co-authors of this paper) were involved in our work. They contributed
%
to the problem definition (\autoref{subsec_ImpactPhysics}),
the bibliographic study (\autoref{sec_relatedWork}),
the data acquisition (\autoref{sec_experimentalData}),
the formalization of the observations, assumptions and hypotheses (\autoref{sec_domainInfo}),
the physics-driven post-processing (\autoref{sec_physics}),
and the interpretation of the case studies (\autoref{sec_caseStudies})}.

\subsection{Validation}

\label{sec_physicalValidation}
\begin{figure}
  \centering
  \includegraphics[width=\linewidth]{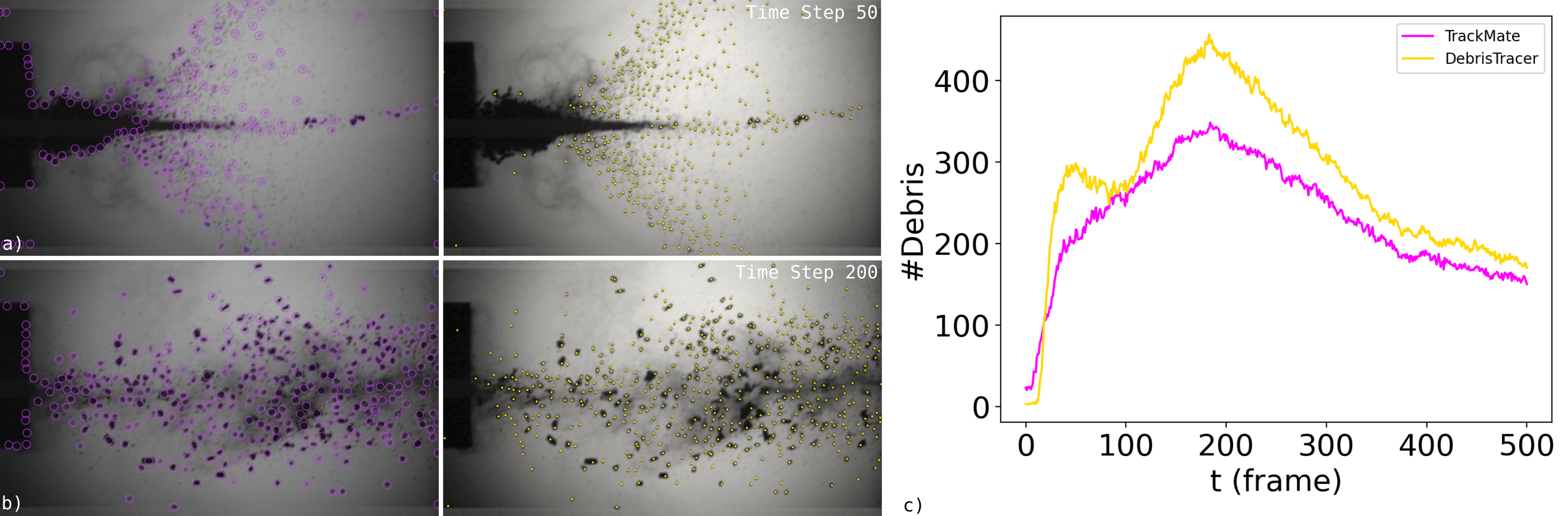}
  \caption{Qualitative and quantitative comparison of debris characterization between
	an established domain tool (\emph{TrackMate} \cite{tinevez2017trackmate, Ershov2022}, purple circles) and \emph{DebrisTracer} (yellow spheres) at time step $50$ (top) and $200$ (bottom) of our reference dataset (\autoref{fig:teaser}). Visually,  our approach makes fewer misses in debris detection \emph{(a)-(b)}. This is confirmed quantitatively  with
  a larger number of identified
  fragments throughout the sequence \emph{(c)}.}
  \label{fig:comparisonDebrisCount}
\end{figure}

We evaluate the accuracy of our approach based on several criteria, including its ability
to predict derived experimental measures (ejected mass and crater profiles, \autoref{sec_applicationContext}), hence assessing its physical relevance.

\noindent
\textbf{\emph{(i)} Debris count:}
\autoref{fig:comparisonDebrisCount} provides a qualitative and quantitative comparison between
\emph{TrackMate} and \emph{DebrisTracer}
in terms of debris identification. As it is based on persistence, \emph{DebrisTracer} manages to correctly identify debris of a variety of sizes, including very small ones
\autoref{fig:comparisonDebrisCount}\emph{(a)-(b)}. This is confirmed quantitatively, \autoref{fig:comparisonDebrisCount}\emph{(c)}, with a larger number of identified debris
fragments
through the entire sequence.

\noindent
\textbf{\emph{(ii)} Trajectory path linearity:}
We evaluate the
improvement of our tracking,
in comparison to the original off-the-shelf topology tracking framework \cite{soler_ldav18}
considered in this work. For that, we focus on the \emph{trajectory paths} (\autoref{sec_physicalTrajectories}) extracted by both approaches (after assignment). Specifically, we evaluate their relevance with regard to the constant velocity assumption (\emph{PA1}, \autoref{sec_domainInfo}). For that, we estimate how well these paths already align with a straight line, by considering the \emph{coefficient of determination}
(in $[0, 1]$)
associated to their linear regression \cite{SeberLee2003, Kalman1960}.
Specifically, the original tracking framework obtains an average coefficient of $0.755$, while our modifications (\autoref{sec_tracking}) yield an average coefficient of $0.910$, a $20.5\%$ improvement.

\noindent
\textbf{\emph{(iii)} Ejected mass estimation:}
For our reference dataset  (\autoref{fig:teaser}), the total ejected mass is evaluated experimentally to $324$ mg \fv{from the crater volume }(\autoref{sec_experimentalData}).
To assess the physical relevance of the tracking frameworks considered in our work, we will use them to
predict
this mass, based on their tracking information.
\emph{TrackMate} does not provide
direct
methods for particle mass estimation \revision{and thus cannot be used directly for this quantitative evaluation. In order to still enable a baseline comparison, we suggest instead the following, easily implementable strategy, based on simple transformations of \emph{TrackMate}'s outputs.}
With the selected detector backend (\autoref{sec_parameterSetting}), each debris is represented in \emph{TrackMate} by a disk of constant
area,
whose size is user parameterized (in our setup, 7-pixel diameter). From this area information, we can estimate the volume of each debris under the hypothesis of spherical debris shape (similarly to \autoref{sec_geometry}), and hence its mass (given the uniform volume density \emph{physical assumption PA3}, \autoref{sec_domainInfo}).
By summing the computed mass of all the identified trajectories, this yields an estimated
total
mass of
\revision{$1,453$ mg}.
This 
\revision{$+348 \%$}
over-estimation can be explained by the fact that
the trajectories provided by \emph{TrackMate}
exhibit many temporal gaps. Then, a single debris
fragment
is tracked by several trajectories at several points in time, resulting in an over-estimation of its contribution to the total ejected mass.
\revision{This issue might be addressed in the future by extending \emph{TrackMate} with a tailored post-processing of trajectories (similar to our approach,
\autoref{sec_physics}).}
In contrast, while
\emph{DebrisTracer} correctly identifies more
fragments (\autoref{fig:comparisonDebrisCount}), it also tracks them individually better over time,
resulting in an
estimated total ejected mass of
$296$ mg. This $-9\%$ under-estimation is much closer to the experimental data (error reduction by a factor of $47.25$), hence assessing the physical relevance
of
our approach.

\begin{figure}
  \centering
  \includegraphics[width=\linewidth]{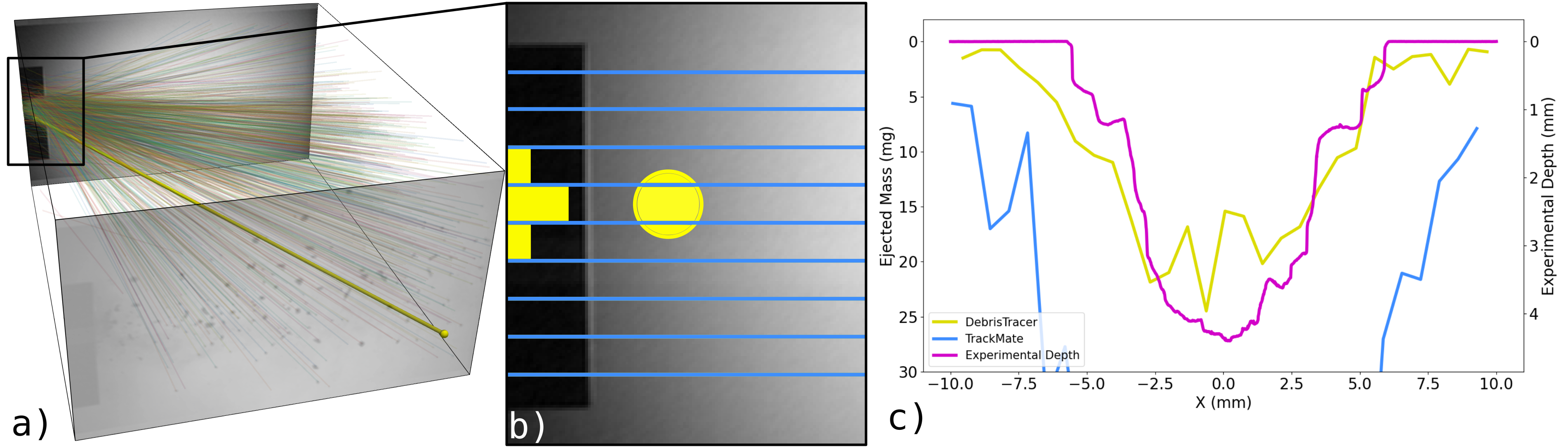}
  \caption{The debris trajectories \emph{(a)} and shapes \emph{(b)} computed by \emph{DebrisTracer} enable the estimation of the distribution of ejected mass along the target \emph{(c)}. This distribution (yellow) nicely coincides with the  experimental crater depth profile (purple), while a similar evaluation
  based on \emph{TrackMate} (blue) yields a global over-estimation of the ejected mass.}
  \label{fig:comparisonCraters}
\end{figure}

\noindent
\textbf{\emph{(iv)} Crater profile reconstruction:}
We further evaluate the physical accuracy of the generated debris trajectories by estimating, from the tracking information, the crater profiles measured experimentally
(\autoref{fig:applicationContext})
on our reference dataset (\autoref{fig:teaser}).
For that, we estimate the ejected mass distribution on the target (i.e., at $X=0$) by binning the target segment on the image, and estimating the contribution of each debris to the bins,
given its computed trajectory and estimated mass, according to
a spherical debris shape hypothesis (\autoref{fig:comparisonCraters}). This mass distribution yields a profile similar in shape to the experimental crater profiles. For visual comparison, we display these profiles in the same reference frame, \autoref{fig:comparisonCraters}\emph{(c)},
by scaling the $Y$-coordinate of the mass distribution according to the estimated
total mass  (c.f., previous paragraph):
the global profile minima are scaled at $91\%$  for \emph{DebrisTracer} and 
\revision{$448\%$}
for \emph{TrackMate},
with regard to
the
global minimum of the crater profile measured experimentally.
\autoref{fig:comparisonCraters} shows that our approach provides an improved estimation of the crater profile, in particular by better capturing the edges of the crater. This validates the physical accuracy of the trajectories and masses estimated by \emph{DebrisTracer}, enabling its reliable use for the interpretation of case studies (\autoref{sec_caseStudies}).

\input{studies.tex}

\subsection{Computational aspects}
The processing of an input dataset by our approach involves several stages, for which we report here the time complexity and the timings obtained on a commodity desktop computer.
First, for each time step, the persistence diagram
is
computed for debris detection
(\autoref{sec_characterization}).
In 2D, this can be
achieved  in $\complexity\big(n_\domain \alpha(n_\domain)\big)$ steps after data sorting, where $n_\domain$  is the number of vertices in $\domain$ and $\alpha$ is the inverse of the Ackermann function. Second,
\revision{persistent maxima are selected in linear time, in at most $\complexity(n_\domain)$ steps. Third,}
persistent maxima are matched between consecutive frames via optimal assignment (\autoref{sec_tracking}), implemented with the Auction algorithm \cite{Bertsekas91} in $\complexity(n_m^2)$ steps in practice, where $n_m \ll n_\domain$ is the number of maxima.
Third, trajectories are refined through physics-based post-processing (\autoref{sec_physics}). Linear regression is applied in linear time (\autoref{sec_physicalTrajectories}), while segment matching
 (\autoref{sec_trajectoryContinuation})
is quadratic in the number of segments in the worst case. Outlier filtering
(\autoref{sec_phyicalLines}) is linear.
Overall,
the three above phases require \theophane{3.5 s.}
of computation for our reference dataset (\autoref{fig:teaser}).
For comparison, \emph{TrackMate}, for a compatible output, took  \theophane{9.7 s.} to process
our reference dataset (\autoref{fig:teaser}), resulting in a \todo{$\times 2.77$} speedup for our approach. To additionally estimate the debris mass distribution (\autoref{sec_geometry}), topological simplification
\revision{($\complexity\big(n_\domain log(n_\domain)\big)$ steps) must be 
performed 
right
before merge tree computation  ($\complexity\big(n_\domain 
\alpha(n_\domain)\big)$ steps), so that the extracted merge-tree leaf regions do 
coincide with the selected persistent maxima. Overall, this mass estimation 
requires another \revision{1.8 s.} of computation with TTK.}

\subsection{Limitations}
In comparison to \emph{TrackMate}, whose setting relies mostly on the adjustment of one important parameter (the estimated debris size), our approach involves more parameters. However, we documented a protocol for their adjustment (\autoref{sec_parameterSetting}) and we showed that the resulting parameters values were stable across datasets, in paricular across modalities (projectile launch versus laser pulse).

Given their considerable speeds with regard to the capture frequency, the early debris are
difficult to track reliably.
Moreover, in certain cases, a non-negligible portion of the early debris blob may leave the image before splitting into multiple connected components, hence preventing the identification of its individual
debris
fragments.
However, this latter issue is
a
limitation of the recording process, which captures in 2D a phenomenon that is intrinsically 3D, hence resulting in inevitable occlusion.
\revision{Also, low persistence maxima are discarded from our analysis and thus, not tracked.
Finally, our debris tracking (\autoref{sec_tracking}) is based on an assignement optimization, which may still produce few incorrect matchings, yielding physically implausible trajectories (later on filtered out by our post-processing, \autoref{sec_physics}), leading to certain fragments not being tracked during certain time intervals.}
Despite this, \emph{DebrisTacer} still manages to identify
\revision{and track}
a sufficient portion of
\revision{debris fragments,}
to enable their reliable
visual
analysis (\autoref{sec_caseStudies}).

Finally, our debris volume estimation could be improved. We currently rely on a spherical debris shape assumption and more accurate estimations might be obtained by considering
advanced shape hypotheses, possibly relying on a pre-documented dictionary of debris shapes.

%
%

%% file: studies.tex
\subsection{\fv{Experimental} case studies}
\label{sec_caseStudies}

To evaluate the application relevance of our
framework, we proceed to its usage for the analysis of
\revision{two}
\fv{dynamic fragmentation experiments.}
\revision{Appendices \ref{sec_laser}, \ref{sec_gallery}}
and our companion video include further datasets, illustrating the stability and versatility of our approach.


\begin{figure}
  \centering
  \includegraphics[width=\linewidth]{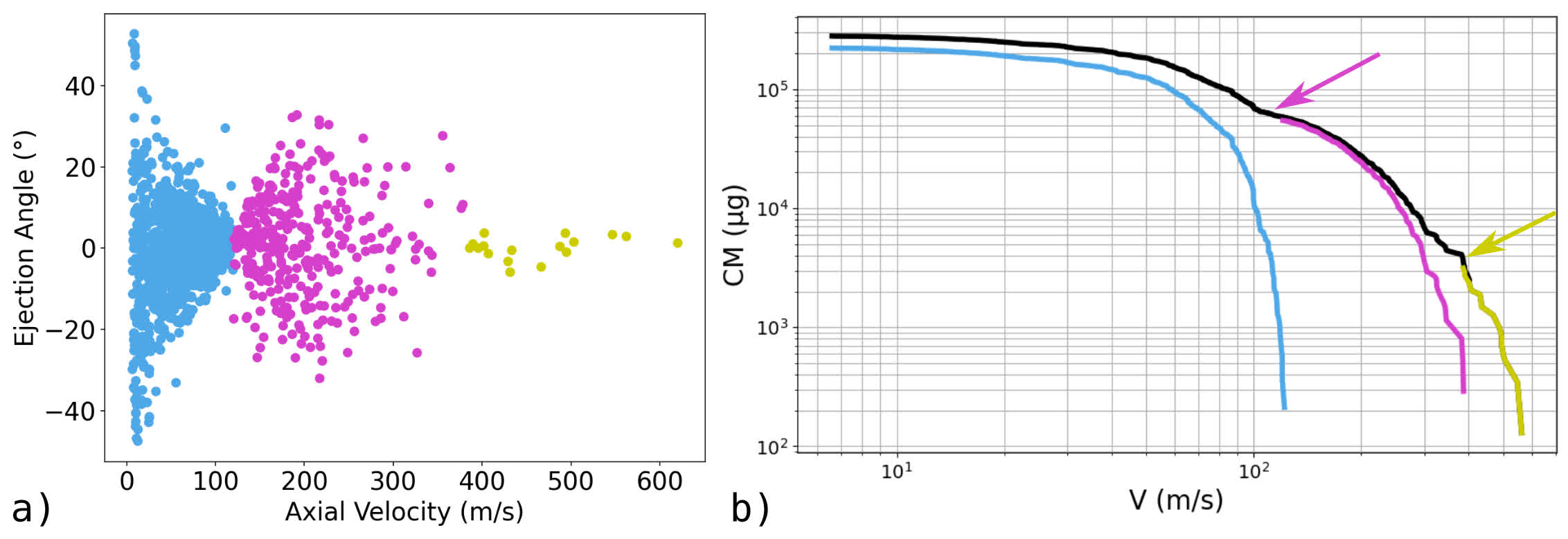}
	\caption{Distributions
	computed by \emph{DebrisTracer} (reference dataset, \autoref{fig:teaser}). The angle/velocity scatterplot \emph{(a)} reveals three distinct modes: thin high-speed (yellow), broad intermediate-speed (purple) and wide slow (blue) ejections.
	This split
	is confirmed by $2$ discontinuities \emph{(b)} in the
	overall
	complementary
	cumulative
	mass/velocity
	distribution (black).}
	\label{fig_teaserStatistics}
\end{figure}

\noindent
\textbf{\emph{(i)} Reference dataset (\autoref{fig:teaser}):}
This use case focuses on
the setup described in \autoref{sec_experimentalData}, with a 90° impact angle against
a
target surface. This dataset is typical of such HVI recordings and we have used it as a reference for setting up the parameters of our approach (\autoref{sec_parameterSetting}).
Our approach enables an accurate tracking of the
debris
fragments
(one color per fragment, \autoref{fig:teaser}),
and
a precise estimation of their shape (enabling their mass estimation, \autoref{sec_geometry}).
Although the initial time steps
reveal a large blob containing most of the debris, \emph{DebrisTracer} still enables a reliable estimation of the debris trajectories throughout the sequence,
based on the linear regression (\autoref{sec_physicalTrajectories}) implementing the \emph{physical assumption PA1} on constant individual velocities (\autoref{sec_domainInfo}).
Then, the identification of debris at a later stage in the sequence still enables the reconstruction of their
early trajectory (thanks to this physical input),
even when most debris overlap (a challenging configuration for traditional topological approaches).
This physics-driven trajectory estimation enables a reliable statistical analysis of the debris trajectories. \autoref{fig_teaserStatistics}\emph{(a)}
shows
the
scatterplot
of
debris ejection angles as a function of their axial velocities. In particular,
this
scatterplot
exhibits a very specific pattern, enabling the visual identification of two axial speed cutoffs, delimiting three distinctive
regimes.
A first regime (yellow) involves
few high-speed debris ($> \fv{400}
~m/s$) within a
thin ejection cone  ($[-10$°$, 10$°$]$). This corroborates the \emph{hypothesis H1} (\autoref{sec_domainInfo}), describing a \fv{multi-phase ejection}.
Specifically, the yellow regime coincides with debris that could be expected from \emph{jetting} (\autoref{subsec_ImpactPhysics}).
A \fv{slower} 
regime,
starts for axial velocities below $\fv{400}
~m/s$. However, unexpectedly, this second regime seems to be
split in two modes (purple and blue). In the first mode (purple), a significant number of intermediate-speed debris
fragments
($[120~m/s, \fv{400}
~m/s]$) travel within an ejection cone of intermediate
width
($[-35$°$, 35$°$]$). In the second mode (blue), a much larger number of
fragments
travel within a larger cone ($[-45$°$, 45$°$]$)
more slowly ($< 120~m/s$), at speeds which
decrease
for increasing ejection angles
(triangle shape, blue point cloud in \autoref{fig_teaserStatistics}\emph{(a)}).
This denotes the energy absorption
capabilities of the target's porous material, characteristic of \emph{spallation} (\autoref{subsec_ImpactPhysics}).
This
corroborates the \emph{hypothesis H2} (\autoref{sec_domainInfo}).
The identification of a first ejection regime (yellow) and the split of the second (purple and blue) is confirmed by the
complementary
cumulative mass distribution, as a function of axial velocity, shown in \autoref{fig_teaserStatistics}\emph{(b)}. In particular, this plot shows clear discontinuities in the global distribution (black), matching the speed cutoffs delimiting the regimes, confirming the split of the debris population into three types (yellow, purple and blue).
Overall,
\emph{DebrisTracer}
confirms
the experts'
hypotheses,
while
providing
new insights \fv{(the second ejection regime exhibits distinct sub-processes)}.

\begin{figure}
  \centering
  \includegraphics[width=\linewidth]{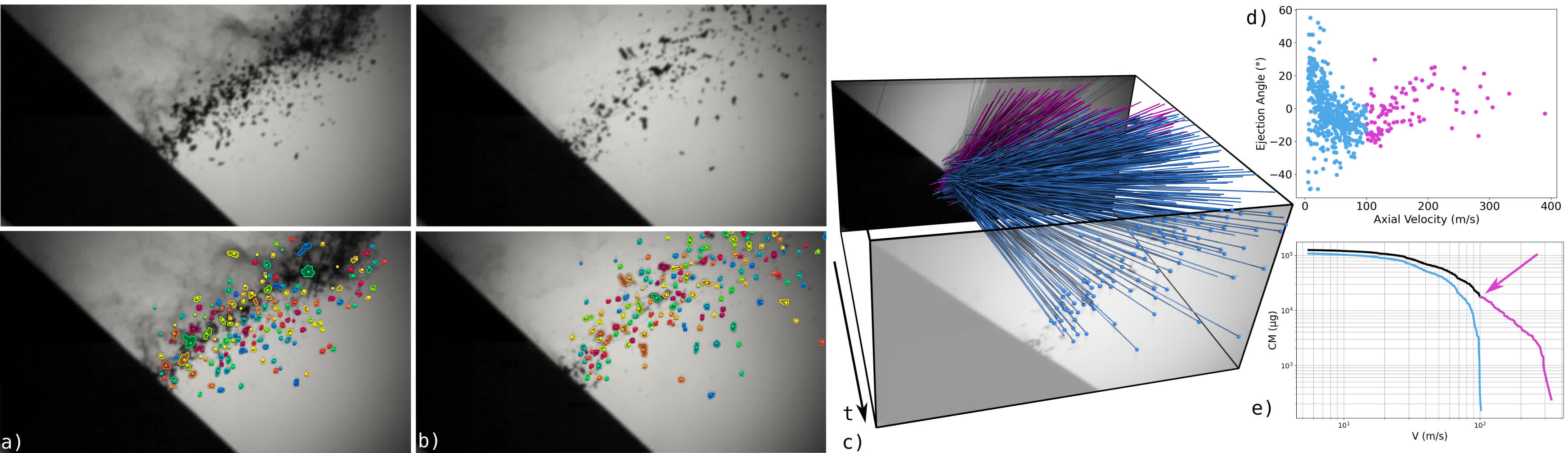}
  \caption{Analyzing an oblique launch (45° impact angle).
  In a pre-process, the dataset is rotated to align the target surface with the vertical axis ($X = 0$).
  In a post-process, the dataset is rotated back to its original angle for visualization purposes.
  Despite the variability in impact angle, \emph{DebrisTracer} still provides an accurate debris trajectory and shape estimation \emph{(a)-(c)}.
  The derived angle/velocity scatterplot \emph{(d)} suggests the presence of
  two
  modes, with a strong bias for the slower debris (blue) towards negative ejection angles (towards the projectile trajectory). This mode split is confirmed visually by
  a
  discontinuity \emph{(e)} in the overall
	complementary
	cumulative mass/velocity distribution (black).}
  \label{fig_45}
\end{figure}

\noindent
\textbf{\emph{(ii)} 45° \fv{projectile} impact (\autoref{fig_45}):} This use case focuses on the same acquisition setup (\autoref{sec_experimentalData}), but this time with a $45$° impact angle.
This experiment shows that the parameters we have selected in \autoref{sec_parameterSetting} are stable across datasets, as the trajectories and the shapes of the debris are also accurately estimated for this dataset. The statistical analysis of the extracted trajectories
also corroborates
the \emph{hypothesis H2} (\autoref{sec_domainInfo}) as the debris with the largest ejection angle are also the slowest. It also corroborates
the \emph{hypothesis H1} (\autoref{sec_domainInfo}), as it reveals
a split
in
two regimes \fv{in this case }(blue and purple),
whose transition is also marked by a discontinuity in the
complementary
cumulative
mass
distribution.
The
scatterplot
of ejection angles as a function of axial speeds shows
a clear distortion of the scatterplot towards negative angles \fv{for debris under $~100~m/s$}.
This
reveals that a non-negligible portion of the debris follow a \emph{reflective ejection}, which is not orthogonal to the surface target, but biased towards the
projectile's trajectory. This original, visual insight  can be
accurately
quantified
thanks to our approach, via statistical moments of the ejection angle (i.e., \theophane{$-2.24$°} on average).

%% file: conclusion.tex
\section{Conclusion}

This application paper presented \emph{DebrisTracer}, a tailored framework for the reliable tracking of debris in HyperVelocity Impact (HVI)
\todo{fast imaging}.
Extensive experiments demonstrated the accuracy improvement provided by \emph{DebrisTracer} over
established tools used by domain experts \cite{tinevez2017trackmate, Ershov2022}.
For instance, \emph{DebrisTracer} reduces by a factor of $47.25$ the error on the
predicted ejected mass.
\revision{Our improved tracking}
enables the trustworthy
visual
analysis of
this complex space-time phenomenon.
Our detailed analyses corroborated the initial hypotheses formulated by domain experts (multi-phase ejection, slow lateral debris), while providing new visual insights which were not directly visible from the experimental recordings. Specifically:
\begin{itemize}
\vspace{-.5ex}
 \item For frontal launches,
 the angle/velocity  distributions
 revealed two \emph{sub-modes} for the slow debris, the slowest mode clearly characterized by decreasing speeds for increasing ejection angles, denoting the energy absorption capabilities of the target material.
 \vspace{-.5ex}
  \item For oblique launches,
  the angle/velocity distributions
  revealed a \emph{partial reflective ejection} towards the projectile's trajectory.
  \vspace{-.5ex}
  \item Without a projectile (laser-based HVI),
  no initial high-speed debris ejection was observed in the early stages of the phenomenon.
\end{itemize}
\vspace{-.5ex}
These finer characterizations of the material's response to the impact enable an improved assessment of its usability in the applications.

We believe our work opens several research avenues for the experimental analysis of HyperVelocity Impacts (HVI). In the future, we will proceed to detailed investigations of additional experimental campaigns, given the new analysis capabilities of \emph{DebrisTracer}. To improve its estimation of ejected mass, and overcome the spherical debris shape assumption, we will consider more advanced strategies, for instance based on shape/mass databases of experimentally collected debris.

%% file: parameterSetting.tex
\section{Parameter setting}
\label{sec_parameterSetting}

This appendix describes our protocol for adjusting the parameters of our approach (\emph{DebrisTracer}) and those of \emph{TrackMate} \cite{tinevez2017trackmate, Ershov2022},
the established tool used by domain experts, to which we will compare.

\noindent
\textbf{\emph{(i)} DebrisTracer:}
The
trajectory post-processing of our approach
(\autoref{sec_physics})  relies on several  thresholds.
However, this step
aims mostly at discarding physical outliers.
Then, these parameters have been set once
and
for all, to conservative values. Specifically,
given two consecutive segments (\autoref{sec_trajectoryContinuation}),
their temporal proximity threshold
($\temporalGap_{max}$) has been set to $30$ time steps,
the maximum absolute value of their angle ($\myangle_{max}$) to $20$°, their spatial continuity ($d_{max}$) to the
\theophane{size of the largest debris}
we have observed (i.e., \theophane{15-pixel diameter}). Finally, we selected a conservative value ($80$°) for the absolute value of the maximum angle $\coneAngle_{max}$ between a trajectory line and the X-axis (\autoref{sec_phyicalLines}).

Earlier in our pipeline,
the tracking of debris
identified
as maxima of $f$ (\autoref{sec_topoTracking}) relies
on the following list of parameters:
\begin{enumerate}
\item
a persistence threshold ($\persistenceThreshold$, \autoref{sec_characterization}),
expressed as a fraction of the function span,
enables,
on a per time step basis,
the identification of the debris as the most salient maxima of $f$.
Since the datasets are provided as gray scale images and since the target is present on all time steps, the function span is stable across time. Hence, this parameter is set to a common value for all time steps.
\item a \emph{scaling vector} ($\scalingVector_x$, $\scalingVector_y$, $\scalingVector_z$, \autoref{sec_tracking}) enables the assignment to favor a horizontal displacement in the debris tracking.
\item a \emph{dedicated destruction weight} ($\destructionCost$, \autoref{sec_tracking}),
expressed as a ratio of the bounding box,
prevents the assignment between consecutive time steps of debris which are too far from each other.
\end{enumerate}

We adjusted these $5$ parameters as follows, by restricting our analysis to the dataset presented in \autoref{fig:teaser}, which is a typical example from our database.
After an initial visual inspection, we identified relevant exploration intervals for these parameters ($\persistenceThreshold \in [0.05, 0.10]$, $\scalingVector_x \in [0, 1]$, $\scalingVector_y \in [0, 1]$, $\scalingVector_z \in [0, 1]$ and $\destructionCost \in [0.01, 0.05]$), which we uniformly sampled ($10$ samples each).
Next, we ran our entire analysis
(Secs. \ref{sec_topoTracking} and \ref{sec_physics})
for each resulting combination of parameter samples and we selected as optimal parameters the
values
which resulted in the largest number of \todo{maintained trajectory lines} for the time step $50$, which is particularly challenging given the small size and large speeds of its debris. Specifically, this exploration yielded the following optimal parameter values:
$\persistenceThreshold = 0.075$,
$\destructionCost = 0.02$,
$\scalingVector_x = 0.1$,
$\scalingVector_y = 0.9$,
$\scalingVector_z = 0.1$.

\noindent
\textbf{\emph{(ii)} TrackMate:} this tool
supports a number of
algorithms for particle detection, all relying on a fundamental parameter, the diameter of the particle to track. For that, we considered
the
average
of the debris sizes
we have observed (i.e., \theophane{7-pixel diameter}). For our experiments, we selected the algorithm \emph{Laplacian of Gaussian}, as it provided the largest number of detections for the time step $50$ of our reference dataset (\autoref{fig:teaser}). Note that this algorithm has also been used in recent papers on debris tracking \cite{ghosh2025quantifying}. Moreover, this approach is a relevant geometrical alternative to our topological strategy (\autoref{sec_characterization}) as it estimates salient maxima, not by persistence, but  via a geometrical characterization (based on a Gaussian smoothing followed by Laplacian computation). Regarding tracking, \emph{TrackMate} supports several approaches, including assignment and overlap based approaches. Specifically, for each method we only maintained the extracted trajectories whose
linear regression was intersecting the target surface at $X=0$. Then, we selected the tracking approach which maximized the number of maintained trajectories at the time step $50$ of our reference dataset (\autoref{fig:teaser}).

%% file: laser.tex
\begin{figure}
  \centering
  \includegraphics[width=\linewidth]{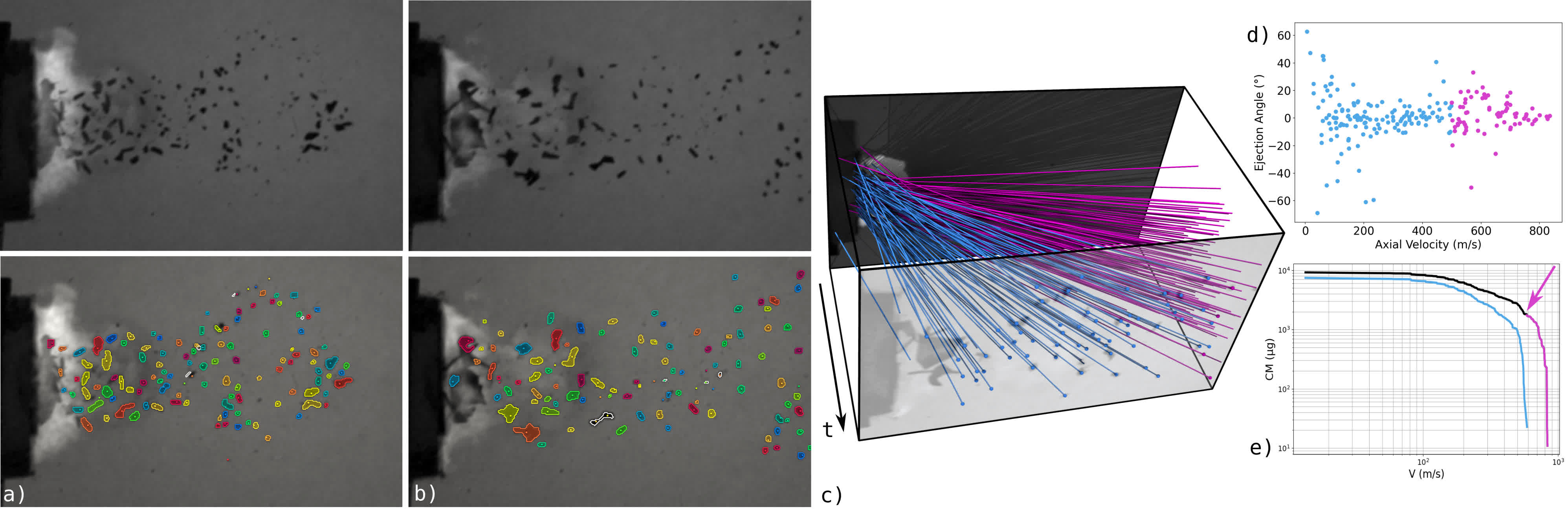}
  \caption{Analyzing a laser-pulse based HVI acquired with ultra-high-speed cameras. Despite this change in physics and acquisition modality, \emph{DebrisTracer} still provides an accurate debris trajectory and shape estimation \emph{(a)-(c)}.
  The derived angle/velocity scatterplot \emph{(d)} also suggests the presence of two groups of debris (purple: fast, blue: slow), also reported in the complementary cumulative mass/velocity distribution \emph{(e)}.
  }
	\label{fig_laser}
\end{figure}

\section{Laser pulse impact}
\label{sec_laser}
%

This use case (\autoref{fig_laser}) covers a different \fv{dynamic fragmentation} modality (based on laser pulse instead of projectile launch) to evaluate the versatility of our approach.
\fv{In this experiment,} specific
cameras
\fv{have been used} (\autoref{sec_experimentalData}), yielding distinct gray scale levels from our
previous use cases
based on projectile launches.
Despite these changes, \emph{DebrisTracer} still identifies accurately the trajectories and the shapes of the debris, in particular the early high-speed debris, given the high capture frequency. This validates the versatility of \emph{DebrisTracer} as well as the stability of its parameters (\autoref{sec_parameterSetting}).
In the early images, \autoref{fig_laser}\emph{(a)}, the debris can be visually separated into two groups, the left one corresponding to slow debris and the right one to faster debris. In \autoref{fig_laser}\emph{(b)}, most of the fast debris have left the image. This observation is also \fv{illustrated}
by the analysis of the
scatterplot
of ejection angles as a function of axial velocity, \autoref{fig_laser}\emph{(c)}, where two groups of debris can be identified, corroborating \emph{hypothesis H1} (\autoref{sec_domainInfo}).
Our
analysis also enables further original insights: in contrast to a projectile-based HVI, the initial high-speed
ejection in a laser-based HVI
does not seem
restricted to a thin cone, as found in our first use case (yellow, \autoref{fig_teaserStatistics}).

%% file: gallery.tex
\section{Result gallery}
\label{sec_gallery}

\begin{figure*}
  \centering
  \includegraphics[width=\linewidth]{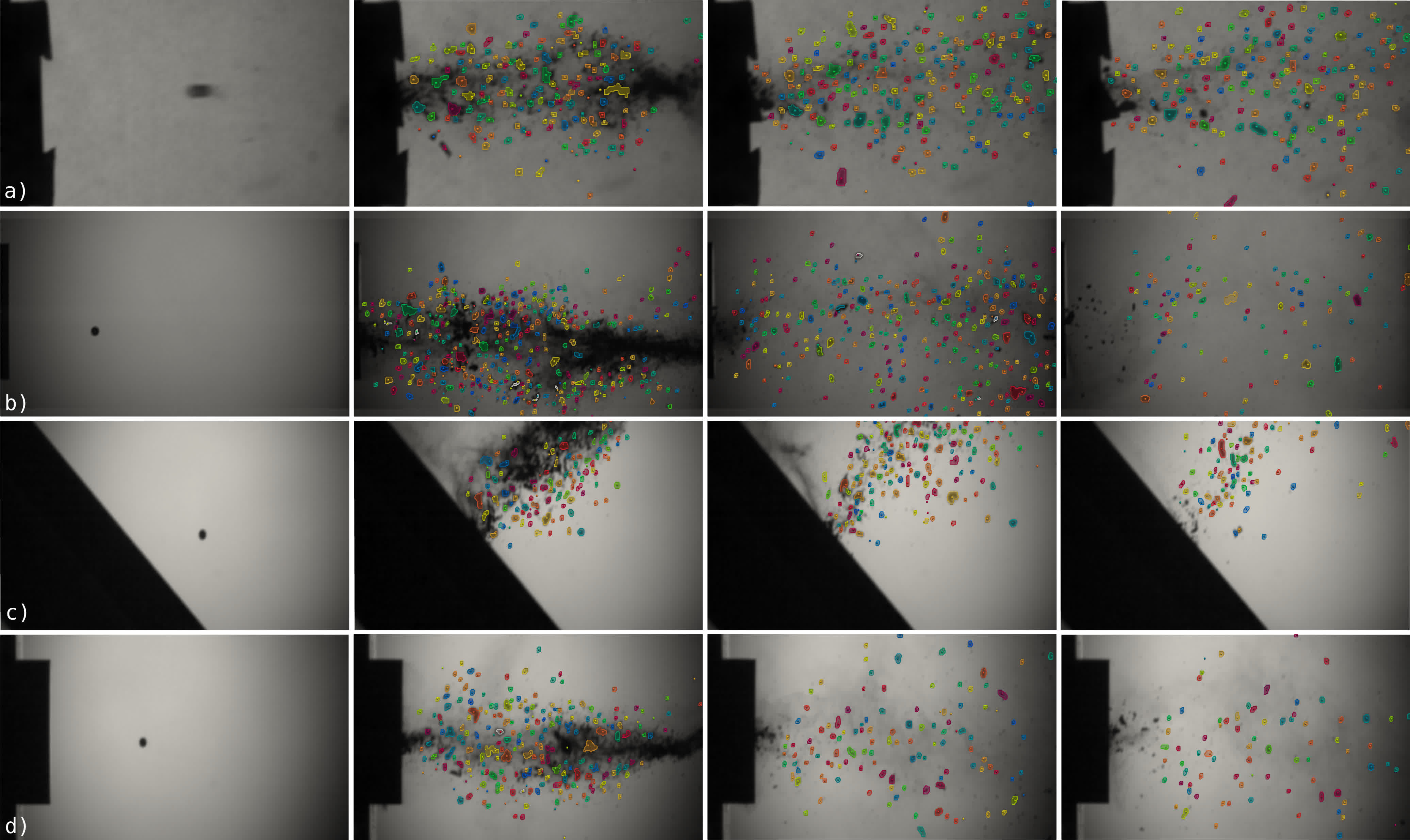}
  \caption{Gallery of tracking results (colored regions) obtained with 
\emph{DebrisTracer} on  several hypervelocity impact acquisitions 
(\emph{(a)}-\emph{(d)}, time is represented with the horizontal axis) based on 
projectile launches (90° and 45° impact angles). \emph{DebrisTracer} enables an 
accurate and physically plausible tracking of debris trajectories, enabling a 
trustworthy statistical analysis of the debris population.}
  \label{fig_gallery}
\end{figure*}

\autoref{fig_gallery} provides a gallery of additional tracking results, obtained with \emph{DebrisTracer}, on several hypervelocity impact acquisitions based on projectile launches (90° and 45° impact angles). These additional results illustrate the versatility and stability of our approach.

